\documentclass[10pt,journal]{IEEEtran}

\newcommand{\contentimage}{x}
\newcommand{\styleimage}{y_{d}}

\newcommand{\stylelabel}{d}
\newcommand{\stylizedimage}{\tilde{y}_{d}}

\newcommand{\contentdomain}{\mathcal{X}}
\newcommand{\styledomain}{\mathcal{Y}_{d}}
\newcommand{\stylelabeldomain}{\mathcal{D}}

\newcommand{\normaldistributionwithparams}{\mathcal{N}(0,I)}

\newcommand{\stylecode}{z}
\newcommand{\imagestylecode}{z_{d}}
\newcommand{\samplestylecode}{z_{s}}
\newcommand{\contentencoder}{E_{c}}
\newcommand{\styleencoder}{E_{s}}

\newcommand{\contentdecoder}{Dec_{c}}
\newcommand{\generator}{G}
\newcommand{\generatorwithparams}{G(\contentimage, \stylecode, \stylelabel)}
\newcommand{\styledis}{D_{s}}
\newcommand{\contentdis}{D_{c}}
\newcommand{\contentcls}{D_{cls}}

\newcommand{\vggfeat}{\mathrm{VGG}_{feat}}
\newcommand{\loss}{\mathcal{L}}
\newcommand{\styleadversarialloss}{\loss_{s\_adv}}
\newcommand{\contentadversarialloss}{\loss_{c\_adv}}
\newcommand{\classificationloss}{\loss_{cls}}
\newcommand{\contentpreservingloss}{\loss_{cp}}
\newcommand{\stylepreservingloss}{\loss_{sp}}
\newcommand{\conditionalidentityloss}{\loss_{cid}}
\newcommand{\fullloss}{\loss_{full}}
\newcommand{\absnorm}{L1}

\newcommand{\betaone}{\beta_{1}}
\newcommand{\betatwo}{\beta_{2}}
\newcommand{\mathenv}[1]{$#1$}
\newcommand{\belongto}[2]{\mathenv{#1\in#2}}
\newcommand{\tabincell}[2]{\begin{tabular}{@{}#1@{}}#2\end{tabular}}  

\newcommand{\nothing}[1]{}
\newcommand{\toremove}[1]{}
\newcommand{\commentout}[1]{}

\newlength{\afterimagevspace}
\PassOptionsToPackage{table}{xcolor}
\usepackage[table]{xcolor}
\usepackage{times}
\usepackage{soul}
\usepackage{url}
\usepackage[utf8]{inputenc}
\usepackage{graphicx}
\usepackage{amsmath}
\usepackage{amsthm}
\usepackage{booktabs}
\usepackage{algorithm}
\usepackage{algorithmic}
\usepackage{amssymb}
\usepackage{ifthen}
\usepackage{caption}
\usepackage{graphicx}
\usepackage{subcaption}
\usepackage{float}
\usepackage{multirow}
\usepackage{color}
\usepackage{microtype}

\begin{document}

\title{Distribution Aligned Multimodal and Multi-Domain Image Stylization}

\author{Minxuan~Lin, Fan~Tang, Weiming~Dong~\IEEEmembership{Member,~IEEE,}, Xiao~Li, Chongyang~Ma, Changsheng~Xu,~\IEEEmembership{Fellow,~IEEE}
\IEEEcompsocitemizethanks{
\IEEEcompsocthanksitem Minxuan Lin, Fan Tang, Weiming Dong and Changsheng Xu are with NLPR, Institute of Automation, Chinese Academy of Sciences, Beijing, China and School of Artificial Intelligence, University of Chinese Academy of Sciences, Beijing, China. E-mail: \{linminxuan2018, tangfan2013, weiming.dong, changsheng.xu\}@ia.ac.cn.
\IEEEcompsocthanksitem Xiao Li is with Microsoft Research Asia. E-mail: pableetoli@gmail.com.
\IEEEcompsocthanksitem Chongyang Ma is with Kuaishou Technology, Beijing, China. E-mail: chongyangma@kuaishou.com.
}
\thanks{}
}

%


\maketitle

\begin{abstract}
Multimodal and multi-domain stylization are two important problems in the field of image style transfer. Currently, there are few methods that can perform both multimodal and multi-domain stylization simultaneously. In this paper, we propose a unified framework for multimodal and multi-domain style transfer with the support of both exemplar-based reference and randomly sampled guidance. The key component of our method is a novel style distribution alignment module that eliminates the explicit distribution gaps between various style domains and reduces the risk of mode collapse. The multimodal diversity is ensured by either guidance from multiple images or random style code, while the multi-domain controllability is directly achieved by using a domain label. We validate our proposed framework on painting style transfer with a variety of different artistic styles and genres. Qualitative and quantitative comparisons with state-of-the-art methods demonstrate that our method can generate high-quality results of multi-domain styles and multimodal instances from reference style guidance or a random sampled style.
\end{abstract}


\section{Introduction}
Style transfer is a typical technique to stylize a content image in the style of another input.
Recently, image-to-image translation methods based on conditional generative adversarial networks~\cite{isola2017image} have played a pivotal role in addressing the problem of style transfer.
While these pioneering techniques have shown promising results for generating a single stylized output from a reference image, two interesting problems have been raised, namely, \textit{multi-domain} and \textit{multimodal} stylization.
Multi-domain stylization methods seek for better \textit{controllability} during the style transfer process, i.e., to generate different styles based on guidance from multiple domains.
Multimodal methods, on the other hand, focus on the \textit{diversity} of generated stylization results, i.e., to synthesize multiple different results which are all consistent with the same reference style.

The majority of existing multi-domain methods, e.g., StarGAN~\cite{choi2018stargan}, are inherited from unpaired conditional image-to-image translation~\cite{zhu2017unpaired}, which \commentout{intrinsically }learns a one-to-one mapping between two domains, and thus loses the ability of synthesizing multimodal results.
Multimodal methods such as MUNIT~\cite{huang2018multimodal}, on the other hand, are usually limited to handling only two domains at one time and do not support multi-domain stylization.
Although several methods~\cite{romero2019smit,yu2019multi} have been proposed to address both multimodal and multi-domain stylization simultaneously, they are restricted to generating images with only random sampling (i.e. generating results by random sampling from the style space) or suffer the issue of mode collapse due to the use of Kullback-Leibler divergence.
\commentout{they are limited in control method diversity, i.e.,} \commentout{they are restricted to generating images with only exemplar guidance (i.e., synthesize stylization results that have the same style with a reference image) or random sampling (i.e. generating results by random sampling from the style space).}

The key of addressing these problems is to construct a style embedding space that (1) preserves style information from reference style images for exemplar-based multimodal stylization; (2) is smooth enough for random sampling based multimodal stylization; (3) has a uniformly covered style distribution for avoiding mode collapse; and (4) provides flexible control using domain labels for multi-domain stylization.

To overcome these challenges, we propose a unified framework that achieves multi-domain and multimodal stylization simultaneously with both exemplar-based guidance and random sample guidance as well as reducing the possibility of mode collapse.
The aligned space has the following properties: (1) a reference style can be extracted from a trained encoder to support exemplar-based stylization; (2) each conditioned space can support multimodal stylization via random sampling; (3) style features sufficient cover the sampling space; and (4) the space is conditioned on domains so that multi-domain stylization is available as well.
We demonstrate the strength of our method on painting style transfer with a variety of artistic styles and genres.
Both qualitative and quantitative comparisons with state-of-the-art methods indicate that our approach can generate high-quality results of multi-domain and multimodal stylization.

\section{Related Work}

\paragraph{Style transfer.}
Gatys et al.~\cite{gatys2016neural} first adopt convolution neural network to deal with a single image stylization problem by an iterative optimization procedure.
For more diversity, several arbitrary style transfer methods are proposed. WCT~\cite{li2017universal} progressively repeats whitening and coloring operations at multiple scales to alter any style patterns.
Huang et al.~\cite{huang2017arbitrary} use the adaptive instance normalization (AdaIN) layer to align with the feature statistics of content and style images.
AvatarNet~\cite{sheng2018avatar} proposes a style decorator to semantically make up the content feature with the style feature in multi-scale layers.
Li et al.~\cite{li2018learning} introduce a transformation matrix to transfer style across different levels flexibly and efficiently.
Sanakoyeu et al.~\cite{sanakoyeu2018style} emphasize the style-aware content constraint to achieve real-time HD style transfer.
Kotovenko et al.~\cite{kotovenko2019content} use a content transformation module to focus on details.
Kotovenko et al.~\cite{kotovenko2019contentandstyle} exchange the content and style of stylized images to disentangle the two elements for better style mix.
Moreover, several attention-aware fashions~\cite{yao2019attention,park2019arbitrary} are proposed, where the models learn to adjust the influencing factor of the style feature for the content feature.
However, none of these above methods support multi-domain style transfer, which are short of the controllability.

\begin{table}
\centering
\resizebox{85mm}{!}{
\begin{tabular}{cccccccc}
\hline
Method  & \tabincell{c}{Unified\\Generator}  & \tabincell{c}{Multiple\\modals} &  \tabincell{c}{Multiple\\domains} & \tabincell{c}{Sample\\Guided} & \tabincell{c}{Exemplar\\Guided} & \tabincell{c}{Feature\\Adaptation} \\
\hline
\hline
\rowcolor[gray]{.9}
Pix2Pix   & $\surd$ & $-$     & $-$     & $-$     & $-$     & $-$     \\
CycleGAN  & $-$     & $-$     & $-$     & $-$     & $-$     & $-$     \\
\rowcolor[gray]{.9}
UNIT      & $-$     & $-$     & $-$     & $-$     & $-$     & $-$     \\
StarGAN   & $\surd$ & $-$     & $\surd$ & $-$     & $-$     & $-$     \\
\rowcolor[gray]{.9}
MUNIT     & $-$     & $\surd$ & $-$     & $\surd$ & $\surd$ & $-$     \\
DRIT      & $-$     & $\surd$ & $-$     & $\surd$ & $\surd$ & $\surd$ \\
\rowcolor[gray]{.9}
UFDN      & $\surd$ & $-$     & $\surd$ & $\surd$ & $-$     & $\surd$ \\
EGSC-IT   & $\surd$ & $\surd$ & $-$     & $-$     & $\surd$ & $-$     \\
\rowcolor[gray]{.9}
DMIT      & $\surd$ & $\surd$ & $\surd$ & $\surd$ & $\surd$     & $-$     \\
SMIT      & $\surd$ & $\surd$ & $\surd$ & $\surd$ & $-$     & $-$     \\
\rowcolor[gray]{.9}
Ours      & $\surd$ & $\surd$ & $\surd$ & $\surd$ & $\surd$ & $\surd$ \\
\hline
\end{tabular}}
\caption{Comparisons with recent methods for image-to-image translation and style transfer.}
\label{tab:comparisons}
\end{table}


\paragraph{Image-to-image translation}
Closely related to style transfer, image-to-image (I2I) translation addresses a more general synthesis problem which shifts the style distribution from one domain to another while maintaining semantic features between images.
CGAN~\cite{mirza2014conditional} renders primitive translation process with a noise condition.
Pix2Pix~\cite{isola2017image} uses conditional generative adversarial networks to transfer images between two domains.
Their methods are further improved by CycleGAN~\cite{zhu2017unpaired} which uses a dual-learning approach and eliminates the requirement of paired data.
While showing promising results, these methods are intrinsically limited to learn a mapping between two domains.

Based on these explorations, several methods attempt to address either multi-domain or multimodal I2I translation.
ACGAN~\cite{odena2017conditional} proposes to append an auxiliary classifier in the discriminator to support multi-domain generation.
For multi-domain translation, ComboGAN~\cite{anoosheh2018combogan} leverages multiple encoder-decoders for altering between different styles.
StarGAN~\cite{choi2018stargan} uses a unified conditional generator for multi-domain synthesis. SGN~\cite{chang2019sym} explores the influence of mixed domains.
For multimodal generation, MSGAN~\cite{mao2019mode} introduces a new constraint which emphasizes on the ratio of the distances between images and their corresponded latent codes.
EGSC-IT~\cite{ma2019exemplar} controls the AdaIN parameters of image generator by a style coding branch.
FUNIT~\cite{liu2019few} deals with multi-domain translation in a few-shot setting.
However, all of them cannot perform multimodal and multi-domain translation simultaneously.

Recently, several methods~\cite{romero2019smit, yu2019multi, viazovetskyi2020stylegan2, liu2020gmm, yang2019multi, lee2020drit++} propose to achieve multi-domain and multimodal synthesis within a single framework.
SMIT~\cite{romero2019smit} uses a combination of random noise and domain condition as guidance for image translation.
DMIT~\cite{yu2019multi} separates content, style, and domain information with different encoders.
However, the limitations of guidance way (only support random sampling) and the difficulty of controlling style space by KL-divergence become their obstacles. 
Concurrent to our work, StarGAN v2~\cite{viazovetskyi2020stylegan2} uses a mapping network to transform a latent code to style code from multiple domains. The multi-branch strategy is also adopted by the discriminator. As a result, the number of parameters will inevitably increase with adding more domains.

\paragraph{Disentangled latent representations.}
Building the mapping between the latent space and the image space promotes the quality and controllability of synthesized output.
VAE~\cite{kingma2013auto} uses the reparametrization tricks to construct the relationships between the two spaces.
CVAE~\cite{sohn2015learning} takes the one-hot label as the condition to construct multiple clusters in the latent space.
AAE~\cite{makhzani2015adversarial} proposes an adversarial strategy to force the latent space distribution to be close to the prior distribution.
UNIT~\cite{liu2017unsupervised} adopts double VAEs to encode the latent vectors into a shared latent space.
MUNIT~\cite{huang2018multimodal} and DRIT~\cite{lee2018diverse} further disentangle the content feature and style feature into disparate manifolds.
To disentangle multi-domain features, UFDN~\cite{liu2018unified} aligns domain representations by an adversarial domain classifier.
Kotovenko et al.~\cite{kotovenko2019contentandstyle} use fixpoint loss to decouple the content and style space.
Similarly, we also encode the two properties by respective encoders.

Table~\ref{tab:comparisons} summarizes different properties of our method and other related techniques.
Most existing methods focus on either multi-domain or multimodal synthesis.
Few of them have explored both with limited support of style guidance.






\section{Our Method}
\label{sec:method}

\begin{figure}
\centering
\includegraphics[width=\linewidth]{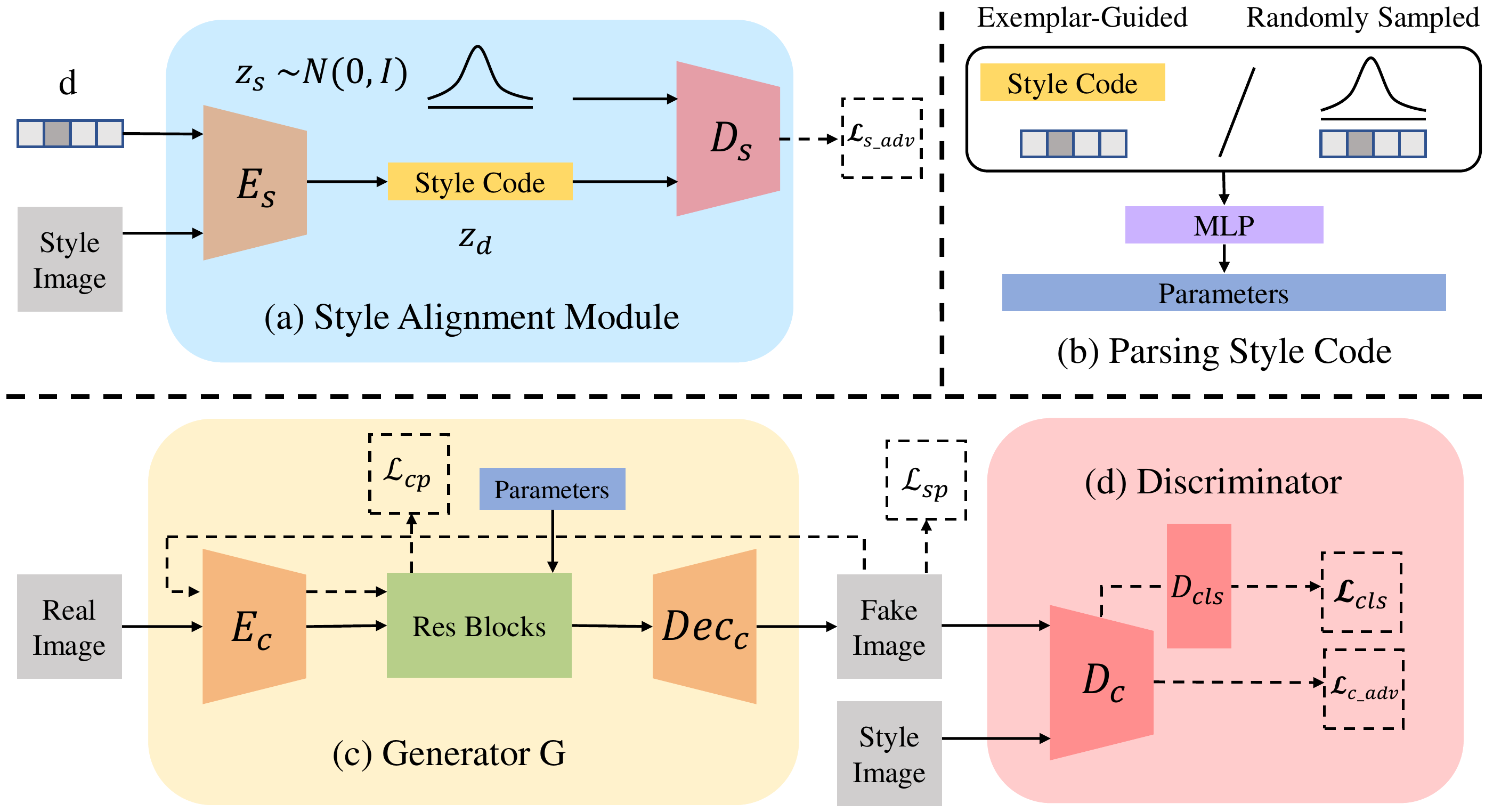}
\caption{Illustration of our entire framework.}
\vspace*{\afterimagevspace}
\label{fig:framework}
\end{figure}

The input of our method includes a natural \emph{content image} \belongto{\contentimage}{\contentdomain} that user wants to stylize, as well as a \emph{style code} $\stylecode$ associated with its domain label \belongto{\stylelabel}{\stylelabeldomain}, i.e., a one-hot vector indicating its style domain.
The style code can be either generated from a reference \emph{style image} \belongto{\styleimage}{\styledomain} of a certain style domain label $\stylelabel$, or directly sampled from a normal Gaussian distribution $\normaldistributionwithparams$.
In case of a style image is provided, the corresponding style code is extracted from our style encoder (Section~\ref{sec:method:style_adv_module}).
The style code $\stylecode$ and the style domain label $\stylelabel$ are then converted to parameters which control the AdaIN layers of our image generator (Section~\ref{sec:method:parsing_code}).
The output image $\stylizedimage$ is finally synthesised by our image generator based on the content image $\contentimage$ (Section~\ref{sec:method:generator}) and the above style information.
Figure~\ref{fig:framework} illustrates our entire framework.


\begin{figure}
\centering
\includegraphics[width=\linewidth]{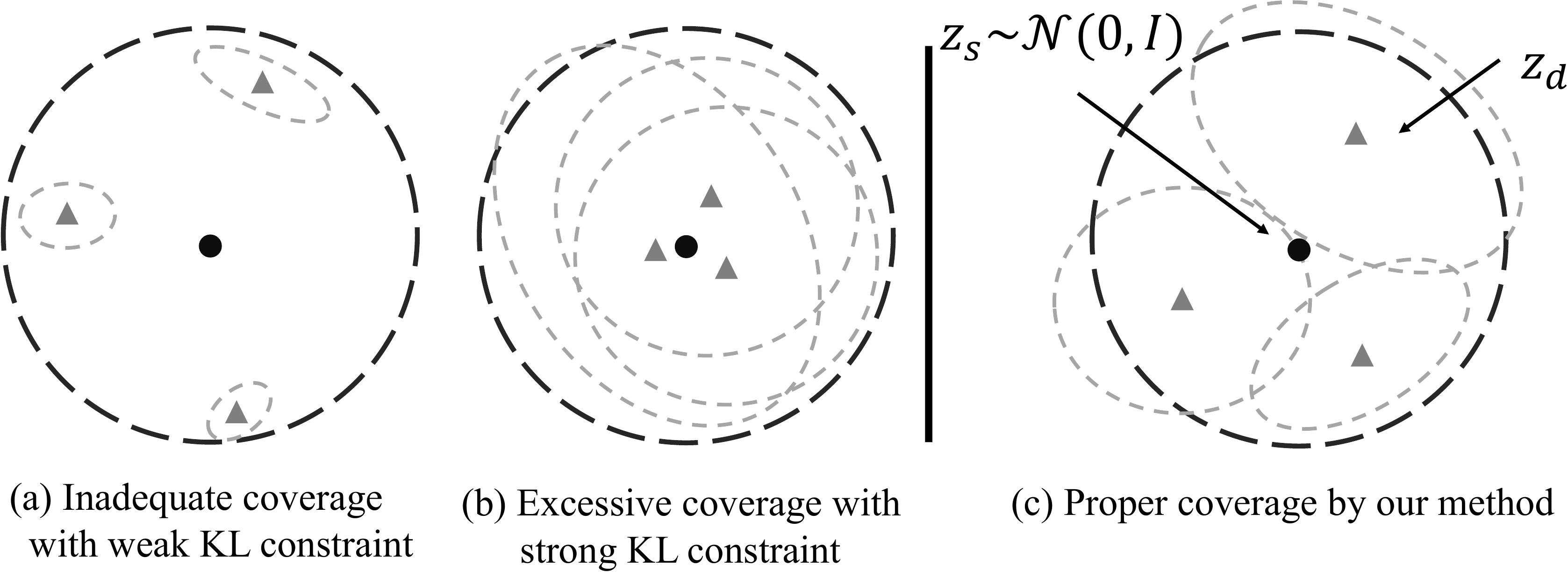}
\caption{The comparison of KL loss and our style alignment module. The improper KL constraint will lead to excessive or inadequate coverage. We avoid this situation by adversarial training to achieve complete and disjoint coverage.}
\label{fig:comparison_kl}
\vspace*{\afterimagevspace}
\end{figure}

\subsection{Style Space Embedding}
The key to integrate multimodal and multi-domain style transfer into a unified framework, without losing ability for either exemplar-guided or random sampling,
is an embedded style space that can be both controlled at the inter-domain level and randomly traversed at the intra-domain level.
In other words, the style space should (1) be clearly separated between different styles via domain label control , and (2) form a smooth space that can be interpolated within a given style domain.
To this end, inspired by CVAE-GAN~\cite{bao2017cvae}, we design a \emph{style alignment module} for style space embedding. 
The embedded style code is then further converted into parameters that control AdaIN layer of our image generation network as in~\cite{huang2017arbitrary}.

\paragraph{Style alignment module.}
\label{sec:method:style_adv_module}
Our style alignment module is an encoder-decoder network\commentout{ similar to~\cite{sohn2015learning}} which constructs the embedded style space from style images of multiple domains.
As shown in Figure ~\ref{fig:framework}(a), we feed a style image $\styleimage$ and its corresponding domain label $\stylelabel$ into the style encoder $\styleencoder$ together to form a one-dimensional style embedding $\imagestylecode$.
The style encoder $\styleencoder$ consists of multiple down-sampling blocks, and global average pooling (GAP) is applied to the final layer\commentout{ of the feature map} to squeeze output style features.

Unlike CVAE-GAN, our style alignment module does not have to accurately reconstruct a given style image.
Instead, the goal of our style alignment module is to eliminate the explicit distribution gaps among various style domains and align them, i.e. different artist styles are controlled by $\stylelabel$, and style space w.r.t each domain are aligned to Gaussian Distribution to enable sampling and smooth interpolation.
Thus, we avoid the reconstruction loss in~\cite{bao2017cvae} since pixel-level reconstruction loss will cause more content-related information encoded into the style code, which obstructs our goal of extracting style information only.

Additionally, the KL-divergence loss used in CVAE-GAN without reconstruction constraint will lead to a trivial solution. \commentout{which the KL-divergence loss under inappropriate weight promptly converges to zero and the style space will collapse.}
Figure~\ref{fig:comparison_kl}(a) and \ref{fig:comparison_kl}(b) show two situations when KL-divergence loss is under an inappropriate weight: (1) A weak KL constraint makes the variance of style feature promptly converges to zero and results in inadequate coverage. (2) A strong KL constraint makes the style feature to be indistinguishable from \mathenv{\normaldistributionwithparams} and results in excessive coverage.
That is, the style space is destroyed in both situations.
Thus, we also remove the KL-divergence term and train our style alignment module with only \emph{style adversarial loss} $\styleadversarialloss$:
\begin{equation}
\begin{split}
\styleadversarialloss&=\mathbb{E}_{z}[\mathrm{log}D_{s}(z_{s})]\\
&+\mathbb{E}_{y_{d},d}[\mathrm{log}(1-D_{s}(E_{s}(y_{d},d)))],
\end{split}
\end{equation}
where $\samplestylecode$ is randomly sampled from a Gaussian distribution $\normaldistributionwithparams$.
The style alignment discriminator $\styledis$ determines whether the unknown style feature points $\stylecode$ are from a Gaussian distribution or generated by $\styleencoder$.

The adversarial loss tends to align the joint distribution of all domain styles (i.e. the unconditioned style space) to a Gaussian distribution.
Consequently, each conditioned space is arranged accordingly to cover different regions, and their union spans the full Gaussian distribution, as illustrated in Figure~\ref{fig:comparison_kl}(c).
A real case of the aligned feature distribution of our trained style space is illustrated in Figure~\ref{fig:distribution}(a) via t-SNE visualization and in Figure~\ref{fig:distribution}(b) via the distance distribution of the L1-distance between random sampled style feature pairs.
Apparently, our style space have a complete and disjoint coverage.

\newcommand\distributionheight{0.36}
\begin{figure}
\centering
\includegraphics[height=\distributionheight\linewidth]{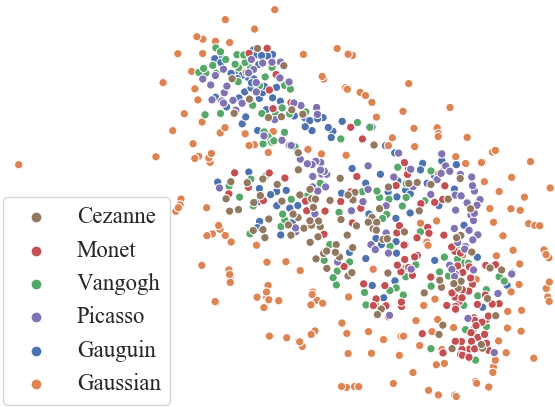}
\includegraphics[height=\distributionheight\linewidth]{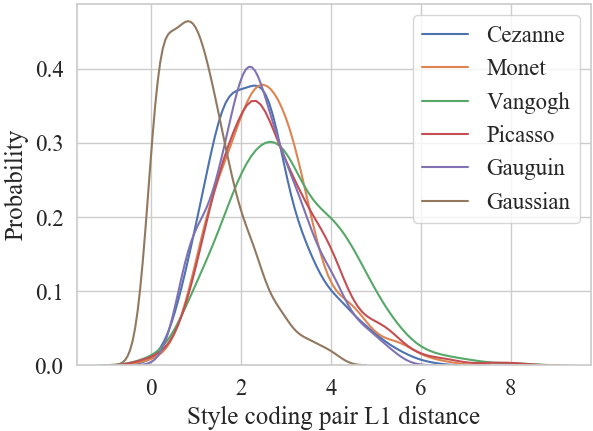}

\vspace{-5pt}
\begin{subfigure}{0.45\linewidth}
    \centering
    \caption{}
\end{subfigure}
\begin{subfigure}{0.45\linewidth}
    \centering
    \caption{}
\end{subfigure}

\vspace{-7pt}
\caption{(a) The t-SNE embedding visualization of style features from different artist domains. (b) The L1 distance distribution of style coding pairs. To show the effect of alignment, two style features from same domain are extracted by $\styleencoder$ as a pair to calculate the distribution of Manhattan distance.}
\label{fig:distribution}
\vspace*{\afterimagevspace}
\end{figure}

\paragraph{Controllable image synthesis.}
\label{sec:method:parsing_code}
Our style alignment module provides two possible ways to parse a style code, i.e., \emph{exemplar guided} and \emph{randomly sampled}.
As shown in Figure~\ref{fig:framework}(b), exemplar-guided style code is extracted from the style alignment encoder, providing precise control information.
And the corresponding stylized result is expected to achieve the same color distribution and texture appearance as the style image $\styleimage$.
Randomly sampled style code enables multimodal stylization in a certain domain via sampling the style code
$\samplestylecode$ and an arbitrary domain label $\stylelabel$.
Similar to~\cite{huang2018multimodal}, we use the style code $\stylecode$ and its corresponding domain label $\stylelabel$ for stylized image generation by controlling the parameters of AdaIN layers.
The style code $\stylecode$ is concatenated with the style label $\stylelabel$ and transformed into channel-wise feature scale and bias for the AdaIN layer by a multi-layer perception network.

\newcommand\guidancecomparewidth{0.0845}
\begin{figure*}
\centering
\begin{subfigure}{\guidancecomparewidth\linewidth}
    \centering
    \includegraphics[width=\linewidth]{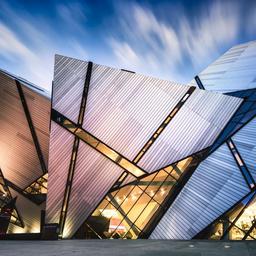}
\end{subfigure}
\begin{subfigure}{\guidancecomparewidth\linewidth}
    \centering
    \includegraphics[width=\linewidth,height=\linewidth]{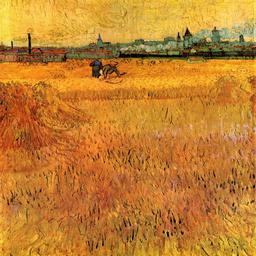}
\end{subfigure}
\begin{subfigure}{\guidancecomparewidth\linewidth}
    \centering
    \includegraphics[width=\linewidth]{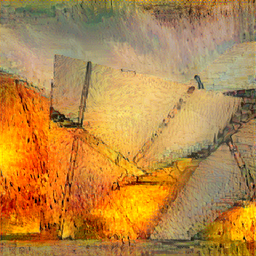}
\end{subfigure}
\begin{subfigure}{\guidancecomparewidth\linewidth}
    \centering
    \includegraphics[width=\linewidth]{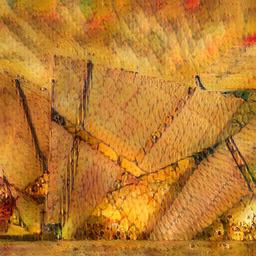}
\end{subfigure}
\begin{subfigure}{\guidancecomparewidth\linewidth}
    \centering
    \includegraphics[width=\linewidth]{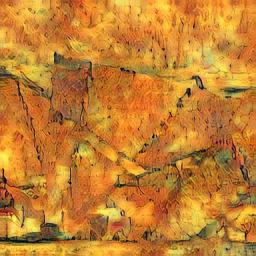}
\end{subfigure}
\begin{subfigure}{\guidancecomparewidth\linewidth}
    \centering
    \includegraphics[width=\linewidth]{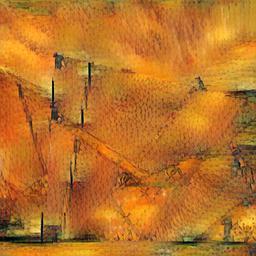}
\end{subfigure}
\begin{subfigure}{\guidancecomparewidth\linewidth}
    \centering
    \includegraphics[width=\linewidth]{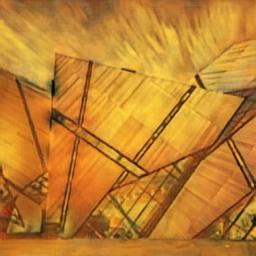}
\end{subfigure}
\begin{subfigure}{\guidancecomparewidth\linewidth}
    \centering
    \includegraphics[width=\linewidth]{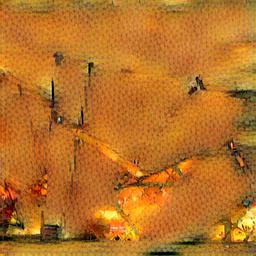}
\end{subfigure}
\begin{subfigure}{\guidancecomparewidth\linewidth}
    \centering
    \includegraphics[width=\linewidth]{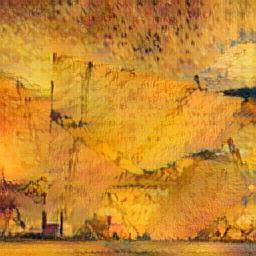}
\end{subfigure}
\begin{subfigure}{\guidancecomparewidth\linewidth}
    \centering
    \includegraphics[width=\linewidth]{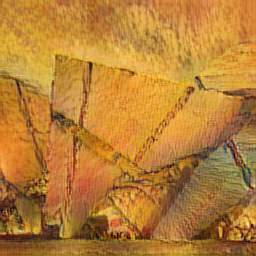}
\end{subfigure}
\begin{subfigure}{\guidancecomparewidth\linewidth}
    \centering
    \includegraphics[width=\linewidth]{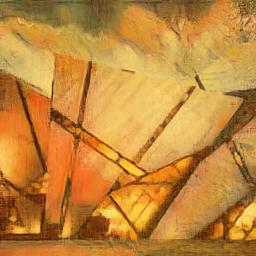}
\end{subfigure}
\begin{subfigure}{\guidancecomparewidth\linewidth}
    \centering
    \includegraphics[width=\linewidth,height=\linewidth]{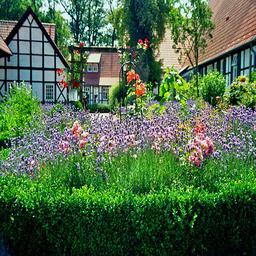}
\end{subfigure}
\begin{subfigure}{\guidancecomparewidth\linewidth}
    \centering
    \includegraphics[width=\linewidth,height=\linewidth]{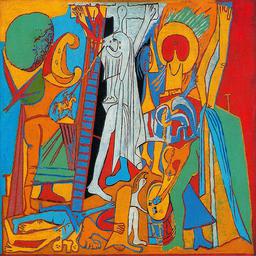}
\end{subfigure}
\begin{subfigure}{\guidancecomparewidth\linewidth}
    \centering
    \includegraphics[width=\linewidth,height=\linewidth]{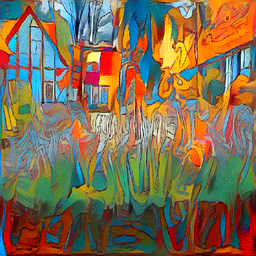}
\end{subfigure}
\begin{subfigure}{\guidancecomparewidth\linewidth}
    \centering
    \includegraphics[width=\linewidth,height=\linewidth]{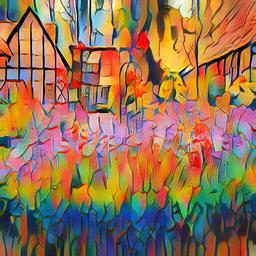}
\end{subfigure}
\begin{subfigure}{\guidancecomparewidth\linewidth}
    \centering
    \includegraphics[width=\linewidth,height=\linewidth]{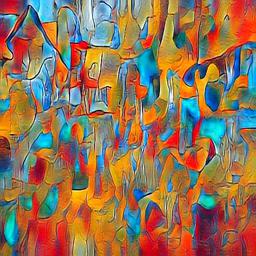}
\end{subfigure}
\begin{subfigure}{\guidancecomparewidth\linewidth}
    \centering
    \includegraphics[width=\linewidth]{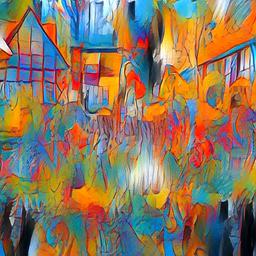}
\end{subfigure}
\begin{subfigure}{\guidancecomparewidth\linewidth}
    \centering
    \includegraphics[width=\linewidth,height=\linewidth]{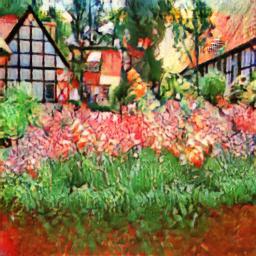}
\end{subfigure}
\begin{subfigure}{\guidancecomparewidth\linewidth}
    \centering
    \includegraphics[width=\linewidth,height=\linewidth]{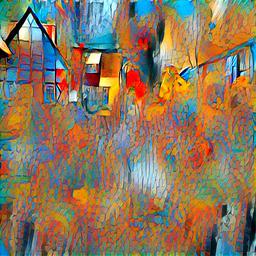}
\end{subfigure}
\begin{subfigure}{\guidancecomparewidth\linewidth}
    \centering
    \includegraphics[width=\linewidth,height=\linewidth]{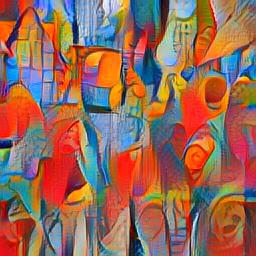}
\end{subfigure}
\begin{subfigure}{\guidancecomparewidth\linewidth}
    \centering
    \includegraphics[width=\linewidth,height=\linewidth]{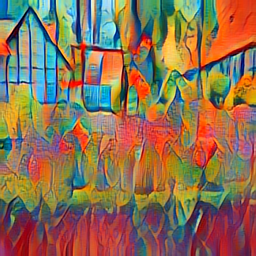}
\end{subfigure}
\begin{subfigure}{\guidancecomparewidth\linewidth}
    \centering
    \includegraphics[width=\linewidth,height=\linewidth]{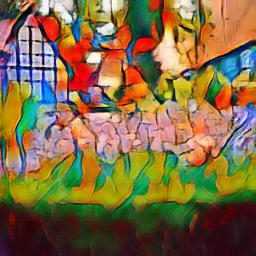}
\end{subfigure}
\begin{subfigure}{\guidancecomparewidth\linewidth}
    \centering
    \includegraphics[width=\linewidth,height=\linewidth]{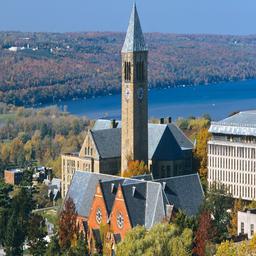}
\end{subfigure}
\begin{subfigure}{\guidancecomparewidth\linewidth}
    \centering
    \includegraphics[width=\linewidth,height=\linewidth]{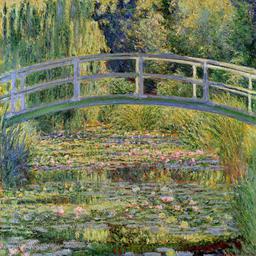}
\end{subfigure}
\begin{subfigure}{\guidancecomparewidth\linewidth}
    \centering
    \includegraphics[width=\linewidth,height=\linewidth]{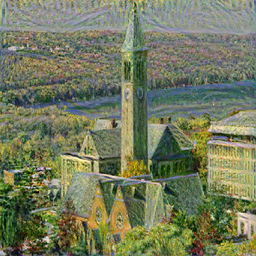}
\end{subfigure}
\begin{subfigure}{\guidancecomparewidth\linewidth}
    \centering
    \includegraphics[width=\linewidth,height=\linewidth]{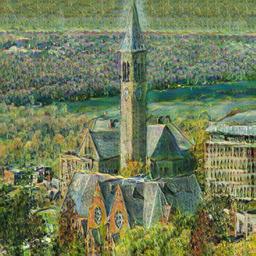}
\end{subfigure}
\begin{subfigure}{\guidancecomparewidth\linewidth}
    \centering
    \includegraphics[width=\linewidth,height=\linewidth]{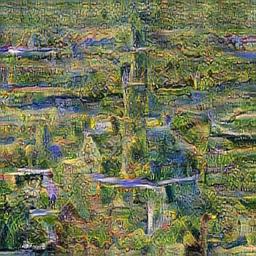}
\end{subfigure}
\begin{subfigure}{\guidancecomparewidth\linewidth}
    \centering
    \includegraphics[width=\linewidth]{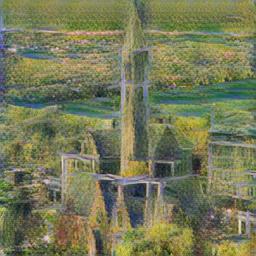}
\end{subfigure}
\begin{subfigure}{\guidancecomparewidth\linewidth}
    \centering
    \includegraphics[width=\linewidth,height=\linewidth]{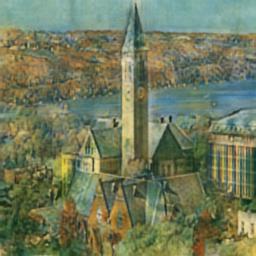}
\end{subfigure}
\begin{subfigure}{\guidancecomparewidth\linewidth}
    \centering
    \includegraphics[width=\linewidth,height=\linewidth]{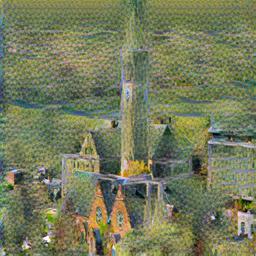}
\end{subfigure}
\begin{subfigure}{\guidancecomparewidth\linewidth}
    \centering
    \includegraphics[width=\linewidth,height=\linewidth]{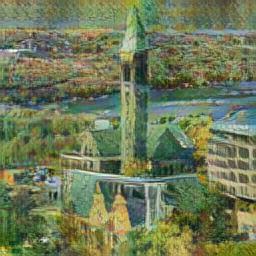}
\end{subfigure}
\begin{subfigure}{\guidancecomparewidth\linewidth}
    \centering
    \includegraphics[width=\linewidth,height=\linewidth]{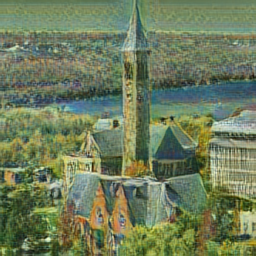}
\end{subfigure}
\begin{subfigure}{\guidancecomparewidth\linewidth}
    \centering
    \includegraphics[width=\linewidth,height=\linewidth]{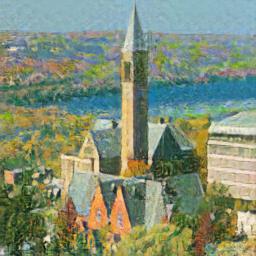}
\end{subfigure}
\begin{subfigure}{\guidancecomparewidth\linewidth}
    \centering
    \includegraphics[width=\linewidth]{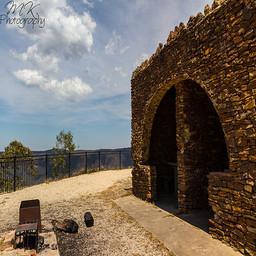}
\end{subfigure}
\begin{subfigure}{\guidancecomparewidth\linewidth}
    \centering
    \includegraphics[width=\linewidth,height=\linewidth]{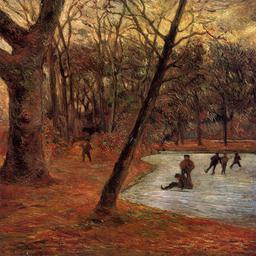}
\end{subfigure}
\begin{subfigure}{\guidancecomparewidth\linewidth}
    \centering
    \includegraphics[width=\linewidth]{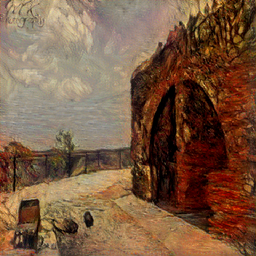}
\end{subfigure}
\begin{subfigure}{\guidancecomparewidth\linewidth}
    \centering
    \includegraphics[width=\linewidth]{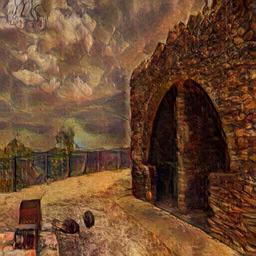}
\end{subfigure}
\begin{subfigure}{\guidancecomparewidth\linewidth}
    \centering
    \includegraphics[width=\linewidth]{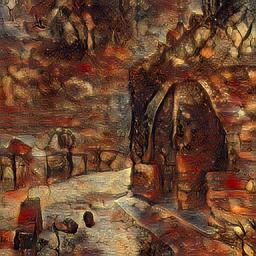}
\end{subfigure}
\begin{subfigure}{\guidancecomparewidth\linewidth}
    \centering
    \includegraphics[width=\linewidth]{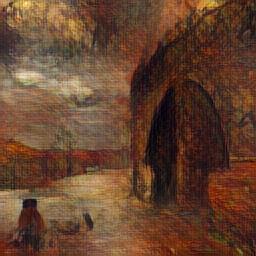}
\end{subfigure}
\begin{subfigure}{\guidancecomparewidth\linewidth}
    \centering
    \includegraphics[width=\linewidth]{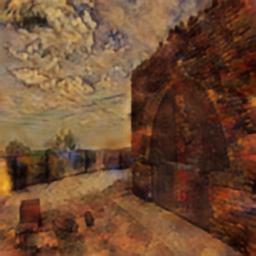}
\end{subfigure}
\begin{subfigure}{\guidancecomparewidth\linewidth}
    \centering
    \includegraphics[width=\linewidth]{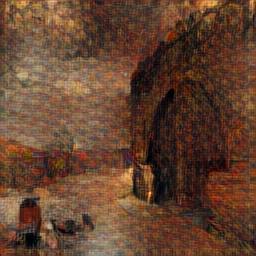}
\end{subfigure}
\begin{subfigure}{\guidancecomparewidth\linewidth}
    \centering
    \includegraphics[width=\linewidth]{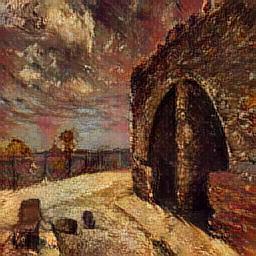}
\end{subfigure}
\begin{subfigure}{\guidancecomparewidth\linewidth}
    \centering
    \includegraphics[width=\linewidth]{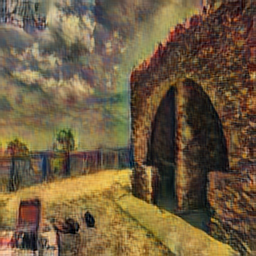}
\end{subfigure}
\begin{subfigure}{\guidancecomparewidth\linewidth}
    \centering
    \includegraphics[width=\linewidth]{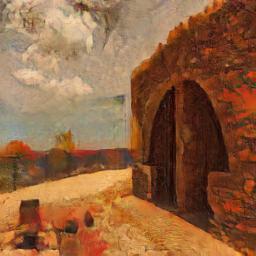}
\end{subfigure}
\begin{subfigure}{\guidancecomparewidth\linewidth}
    \centering
    \includegraphics[width=\linewidth]{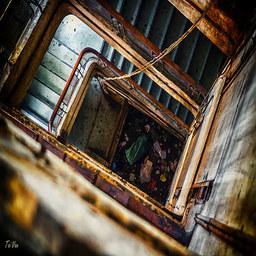}
\end{subfigure}
\begin{subfigure}{\guidancecomparewidth\linewidth}
    \centering
    \includegraphics[width=\linewidth,height=\linewidth]{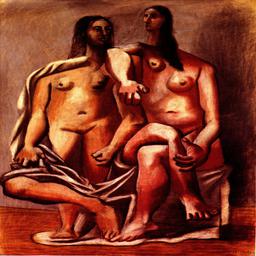}
\end{subfigure}
\begin{subfigure}{\guidancecomparewidth\linewidth}
    \centering
    \includegraphics[width=\linewidth]{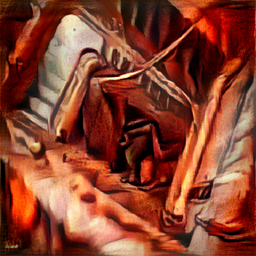}
\end{subfigure}
\begin{subfigure}{\guidancecomparewidth\linewidth}
    \centering
    \includegraphics[width=\linewidth]{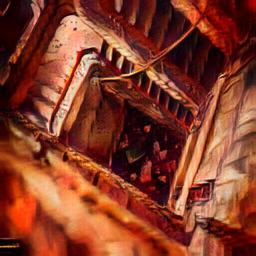}
\end{subfigure}
\begin{subfigure}{\guidancecomparewidth\linewidth}
    \centering
    \includegraphics[width=\linewidth]{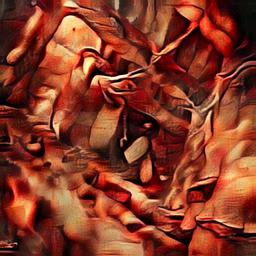}
\end{subfigure}
\begin{subfigure}{\guidancecomparewidth\linewidth}
    \centering
    \includegraphics[width=\linewidth]{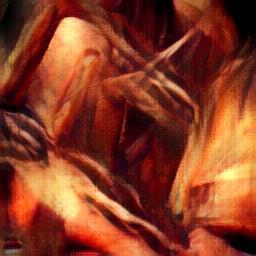}
\end{subfigure}
\begin{subfigure}{\guidancecomparewidth\linewidth}
    \centering
    \includegraphics[width=\linewidth]{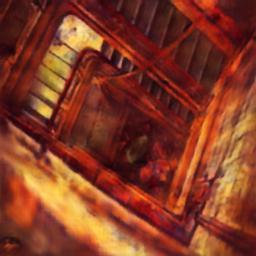}
\end{subfigure}
\begin{subfigure}{\guidancecomparewidth\linewidth}
    \centering
    \includegraphics[width=\linewidth]{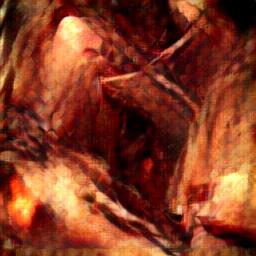}
\end{subfigure}
\begin{subfigure}{\guidancecomparewidth\linewidth}
    \centering
    \includegraphics[width=\linewidth]{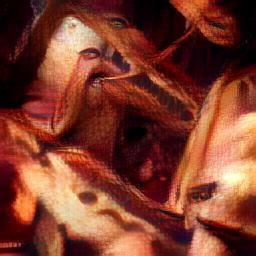}
\end{subfigure}
\begin{subfigure}{\guidancecomparewidth\linewidth}
    \centering
    \includegraphics[width=\linewidth]{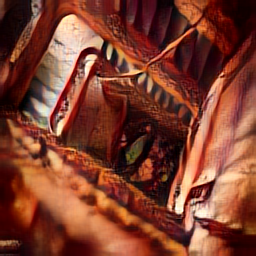}
\end{subfigure}
\begin{subfigure}{\guidancecomparewidth\linewidth}
    \centering
    \includegraphics[width=\linewidth]{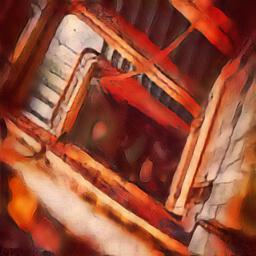}
\end{subfigure}

\begin{subfigure}{\guidancecomparewidth\linewidth}
    \centering
    \caption{\footnotesize Photo}
\end{subfigure}
\begin{subfigure}{\guidancecomparewidth\linewidth}
    \centering
    \caption{\footnotesize Exemplar}
\end{subfigure}
\begin{subfigure}{\guidancecomparewidth\linewidth}
    \centering
    \caption{\footnotesize Gatys}
\end{subfigure}
\begin{subfigure}{\guidancecomparewidth\linewidth}
    \centering
    \caption{\footnotesize AdaIN}
\end{subfigure}
\begin{subfigure}{\guidancecomparewidth\linewidth}
    \centering
    \caption{\footnotesize WCT}
\end{subfigure}
\begin{subfigure}{\guidancecomparewidth\linewidth}
    \centering
    \caption{\footnotesize AvatarNet}
\end{subfigure}
\begin{subfigure}{\guidancecomparewidth\linewidth}
    \centering
    \caption{\footnotesize MUNIT}
\end{subfigure}
\begin{subfigure}{\guidancecomparewidth\linewidth}
    \centering
    \caption{\footnotesize AAMS}
\end{subfigure}
\begin{subfigure}{\guidancecomparewidth\linewidth}
    \centering
    \caption{\footnotesize SANet}
\end{subfigure}
\begin{subfigure}{\guidancecomparewidth\linewidth}
    \centering
    \caption{\footnotesize LinearST}
\end{subfigure}
\begin{subfigure}{\guidancecomparewidth\linewidth}
    \centering
    \caption{\footnotesize Ours}
\end{subfigure}

\caption{Exemplar-guided stylization results from different methods. The content and style images are shown in the left two columns. The remaining columns demonstrate output images generated by several popular style transfer approaches and our method.}
\label{fig:guidance_comparison}
\vspace*{\afterimagevspace}
\end{figure*}

\subsection{Stylized Image Generation}
\label{sec:method:generator}
\paragraph{Image generator.}
Given a content image $\contentimage$ and controlling parameters from $\stylecode$ and $\stylelabel$, the output image $\stylizedimage$ is synthesized by the stylized image generator $\generator$.
Figure~\ref{fig:framework}(c) illustrates our generator framework.
Inspired by CycleGAN~\cite{zhu2017unpaired}, our network is constructed by an encoder-decoder architecture which contains several down-sampling layers, residual blocks and up-sampling layers.
Different from other image-to-image translation and stylization methods~\cite{huang2018multimodal,lee2018diverse} which use consistent normalization methods for most layers, we employ Instance Normalization (IN)~\cite{ulyanov2016instance} in down-sampling layers and first half of residual blocks, AdaIN in second half of residual blocks and Layer Normalization (LN)~\cite{ba2016layer} in up-sampling layers to avoid irregular artifacts.

\paragraph{Loss functions.}
The goal of our stylized image generator network is to generate images that both preserve fidelity with the original content image $\contentimage$ and consistency with the style code $\stylecode$.
In order to ensure the stylized output image preserves the semantic content of the content image $\contentimage$, we use a \emph{content preserving loss} which constrains the output stylized image to achieve same encoded content feature as input content image:
\begin{equation}
\contentpreservingloss=\lVert E_c(G(\contentimage, \stylecode, \stylelabel))-E_c(\contentimage) \rVert_1,
\end{equation}
where the output $\generatorwithparams$ is a stylized image $\stylizedimage$. $\absnorm$ distance is used as the metric.

To ensure the consistency between style image $\styleimage$ and the synthesis output $\stylizedimage$ during the training process, we apply a \emph{style preserving loss}\commentout{ similar to \cite{johnson2016perceptual}}, which computes the $\absnorm$ distance of the gram matrix on multi-scale feature layers of a pre-trained VGG-16 classification network:
\begin{equation}
\stylepreservingloss=\sum_{i=1}^{4}\lVert \vggfeat^{i}(\styleimage)-\vggfeat^{i}(\stylizedimage) \rVert_1,
\end{equation}
where \mathenv{\vggfeat^{i}(\cdot)} indicates the gram matrix of the $i$-th feature map of the VGG network pretrained on ImageNet.

Furthermore, we introduce a \emph{conditional identity loss} to preserve the content fidelity without affecting the output quality.
Specifically, we constrain an identity mapping when using same style image both as style  and content input:
\begin{equation}
\conditionalidentityloss=\lVert \contentdecoder(E_c(\styleimage), E_s(\styleimage, \stylelabel), \stylelabel)-\styleimage \rVert_1,
\end{equation}
where $\contentencoder$ and $\styleencoder$ encode features from the content image and the style image, respectively. Conditioned by the style label $\stylelabel$, $\contentdecoder$ reconstructs the style image $\styleimage$ under $\absnorm$ metric.

To generate realistic results, we use multi-scale patch-based discriminators for adversarial training and auxiliary classifiers for domain classification in Figure~\ref{fig:framework}(d), similar to \cite{choi2018stargan}:
\begin{equation}
\begin{split}
\contentadversarialloss&=\mathbb{E}_{\styleimage}[\mathrm{log}D_{c}(\styleimage)]\\
&+\mathbb{E}_{\contentimage, \styleimage, \stylelabel}[\mathrm{log}(1-D_{c}(\generator(\contentimage, \styleimage, \stylelabel)))],
\end{split}
\end{equation}
where \mathenv{\contentdis} is the multi-scale discriminator network. LSGAN~\cite{mao2017least} loss is used for adversarial training.
The auxiliary classification loss is applied to constrain the output stylized image $\stylizedimage$ and input style guidance $\styleimage$ into the same style domain:
\begin{equation}
\begin{split}
\classificationloss&=\mathbb{E}_{\styleimage}[\mathcal{L}_{D_{cls}}(d|\styleimage)]\\
&+\mathbb{E}_{\contentimage, \styleimage, \stylelabel}[\mathcal{L}_{D_{cls}}(d|G(\contentimage, \styleimage, \stylelabel))],
\end{split}
\end{equation}
where $\loss_{\contentcls}$ is the cross-entropy loss.

The final objective for our generator, discriminator, style alignment module and auxiliary classifier is formulated as:
\begin{equation}
\begin{split}
\min\limits_{E_{s},G,D_{cls}}\max\limits_{D_{s},D_{c}}\fullloss&=\lambda_{1}\contentadversarialloss+\lambda_{2}\styleadversarialloss\\&+\lambda_{3}(\conditionalidentityloss+\contentpreservingloss)+\lambda_{4}\stylepreservingloss+\lambda_{5}\classificationloss.
\end{split}
\end{equation}
where $\{ \lambda_i \}$ denotes the relative importance among these objectives.
We set \mathenv{\lambda_{1}, \lambda_{5}=1}, \mathenv{\lambda_{2}=10}, \mathenv{\lambda_{3}=100}, \mathenv{\lambda_{4}=30}.

\section{Experiments}

\subsection{Experimental Setup}

\paragraph{Implementation details.}
We implement the proposed framework using PyTorch\commentout{~\cite{paszke2017automatic}}.
The input resolution to our network is $256\times256$.
We set the dimension of style code to $20$.
For network training, we use Gaussian weight initialization and Adam~\cite{kingma2014adam} optimizer.
The learning rate, $\betaone$ and $\betatwo$ are set to $2e-4$, $0.5$, and $0.999$, respectively.
The full model is trained with $350,000$ iterations.

\newcommand\randomsamplingwidth{0.09035}
\newcommand\randomsamplingwidthwider{0.295}
\begin{figure*}
\centering
\begin{subfigure}{\randomsamplingwidth\linewidth}
    \includegraphics[width=\linewidth]{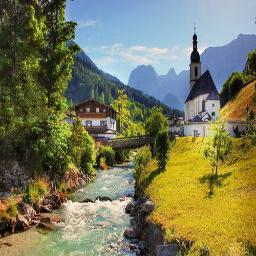}
    \includegraphics[width=\linewidth]{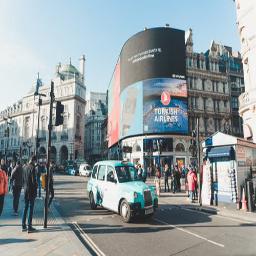}
    \includegraphics[width=\linewidth]{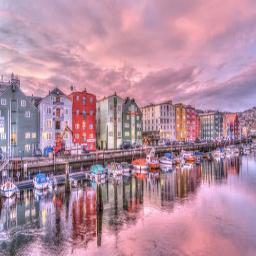}
    \includegraphics[width=\linewidth]{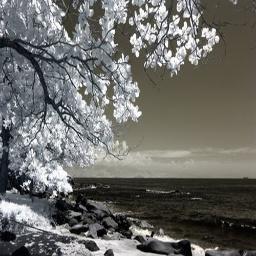}
\end{subfigure}
\begin{subfigure}{2pt}
    \rule{1pt}{195pt}%
\end{subfigure}
\begin{subfigure}{\randomsamplingwidth\linewidth}
    \includegraphics[width=\linewidth]{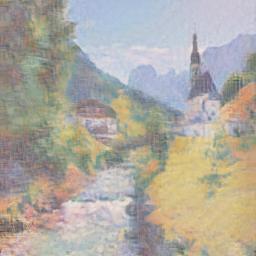}
    \includegraphics[width=\linewidth]{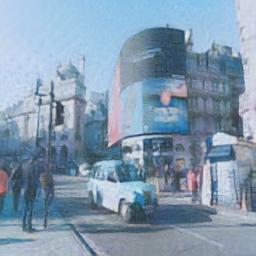}
    \includegraphics[width=\linewidth]{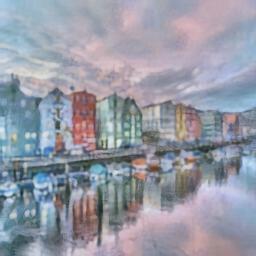}
    \includegraphics[width=\linewidth]{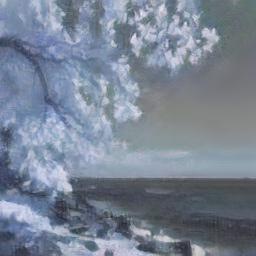}
\end{subfigure}
\begin{subfigure}{\randomsamplingwidth\linewidth}
    \includegraphics[width=\linewidth]{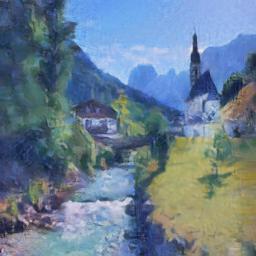}
    \includegraphics[width=\linewidth]{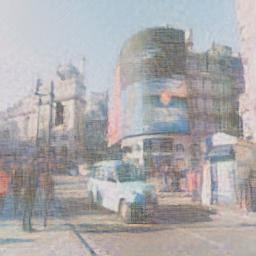}
    \includegraphics[width=\linewidth]{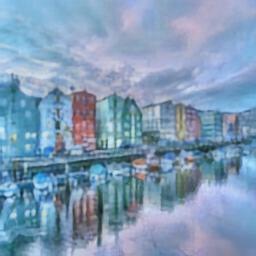}
    \includegraphics[width=\linewidth]{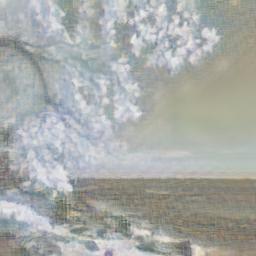}
\end{subfigure}
\begin{subfigure}{\randomsamplingwidth\linewidth}
    \includegraphics[width=\linewidth]{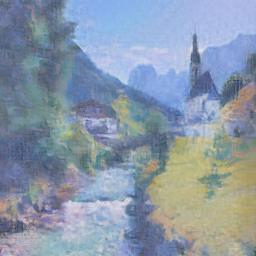}
    \includegraphics[width=\linewidth]{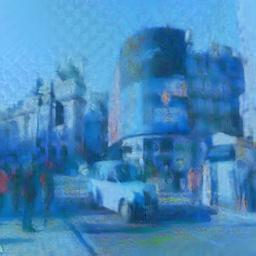}
    \includegraphics[width=\linewidth]{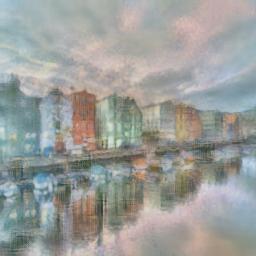}
    \includegraphics[width=\linewidth]{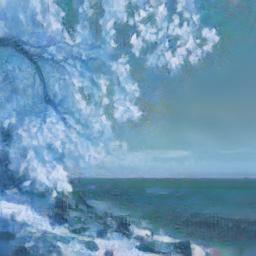}
\end{subfigure}
\begin{subfigure}{2pt}
    \rule{1pt}{195pt}%
\end{subfigure}
\begin{subfigure}{\randomsamplingwidth\linewidth}
    \includegraphics[width=\linewidth]{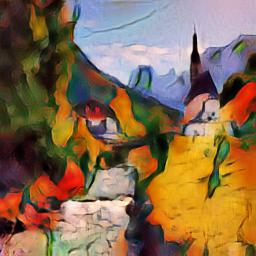}
    \includegraphics[width=\linewidth]{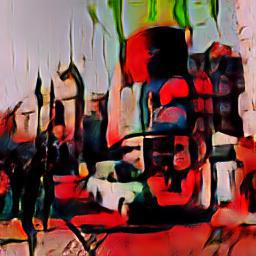}
    \includegraphics[width=\linewidth]{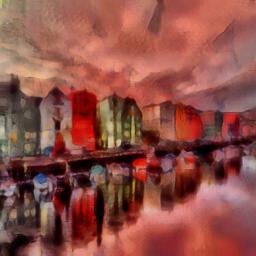}
    \includegraphics[width=\linewidth]{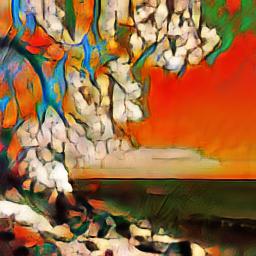}
\end{subfigure}
\begin{subfigure}{\randomsamplingwidth\linewidth}
    \includegraphics[width=\linewidth]{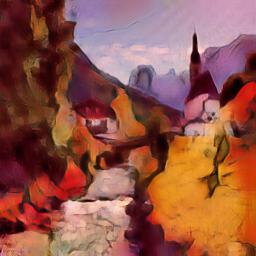}
    \includegraphics[width=\linewidth]{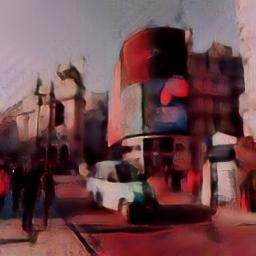}
    \includegraphics[width=\linewidth]{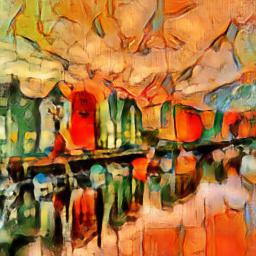}
    \includegraphics[width=\linewidth]{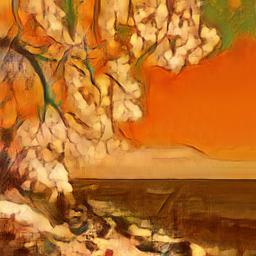}
\end{subfigure}
\begin{subfigure}{\randomsamplingwidth\linewidth}
    \includegraphics[width=\linewidth]{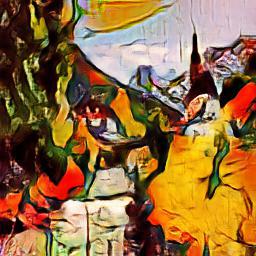}
    \includegraphics[width=\linewidth]{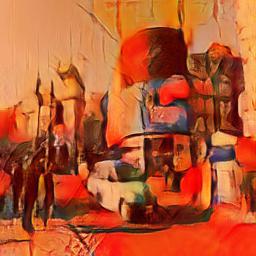}
    \includegraphics[width=\linewidth]{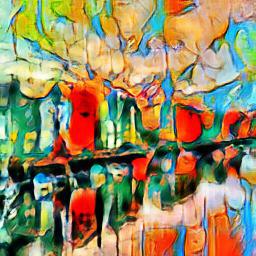}
    \includegraphics[width=\linewidth]{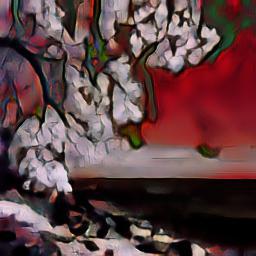}
\end{subfigure}
\begin{subfigure}{2pt}
    \rule{1pt}{195pt}%
\end{subfigure}
\begin{subfigure}{\randomsamplingwidth\linewidth}
    \includegraphics[width=\linewidth]{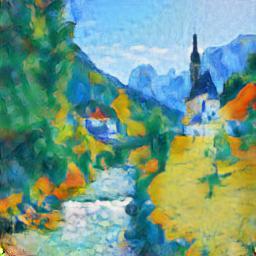}
    \includegraphics[width=\linewidth]{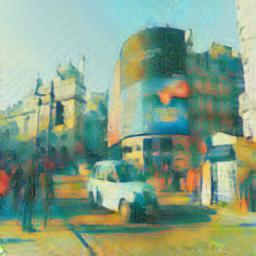}
    \includegraphics[width=\linewidth]{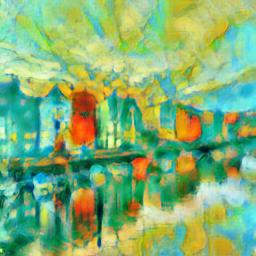}
    \includegraphics[width=\linewidth]{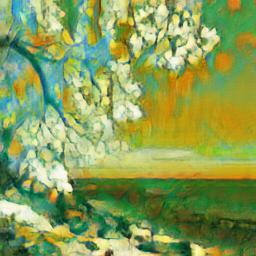}
\end{subfigure}
\begin{subfigure}{\randomsamplingwidth\linewidth}
    \includegraphics[width=\linewidth]{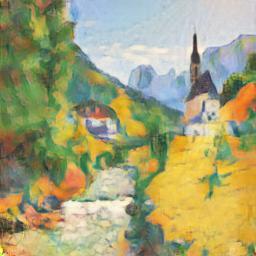}
    \includegraphics[width=\linewidth]{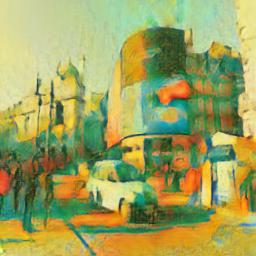}
    \includegraphics[width=\linewidth]{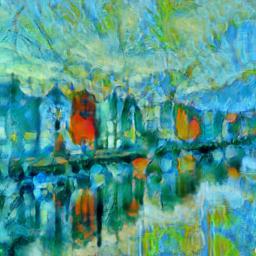}
    \includegraphics[width=\linewidth]{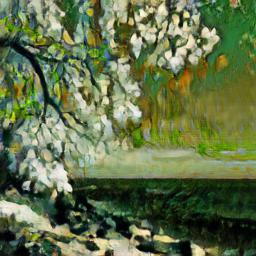}
\end{subfigure}
\begin{subfigure}{\randomsamplingwidth\linewidth}
    \includegraphics[width=\linewidth]{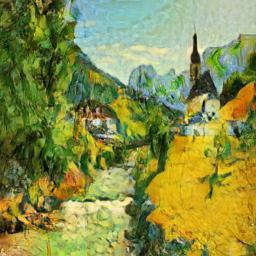}
    \includegraphics[width=\linewidth]{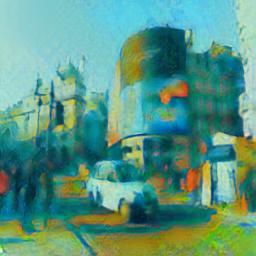}
    \includegraphics[width=\linewidth]{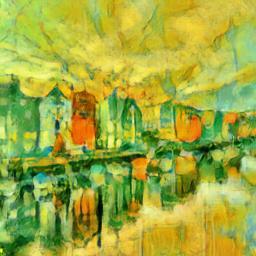}
    \includegraphics[width=\linewidth]{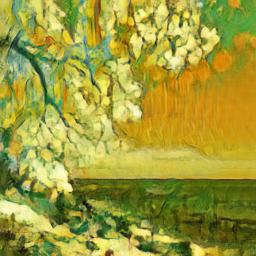}
\end{subfigure}

\begin{subfigure}{0.085\linewidth}
    \centering
    \caption{Photo}
\end{subfigure}
\begin{subfigure}{\randomsamplingwidthwider\linewidth}
    \centering
    \caption{Monet}
\end{subfigure}
\begin{subfigure}{\randomsamplingwidthwider\linewidth}
    \centering
    \caption{Picasso}
\end{subfigure}
\begin{subfigure}{\randomsamplingwidthwider\linewidth}
    \centering
    \caption{C\'ezanne}
\end{subfigure}

\caption{Random sampling guided stylization results of different domains (artists). Random style codes are sampled from a standard normal distribution. (a): Content images; (b)-(d): Our stylization results using the styles of three artists.}
\label{fig:sample_comparison}
\vspace*{\afterimagevspace}
\end{figure*}

\paragraph{Training and test data.}
For multi-domain training, we collect a total of $1303$ paintings from five artists on \textit{Wikiart}\footnote{\url{https://www.wikiart.org/}}, including Monet (458), Van Gogh (184), C$\acute{e}$zanne (257), Gauguin (245), and Picasso (159) as style reference images. Each artist corresponds to one style domain.
The content image sets for training are from the photo2art dataset of CycleGAN\commentout{~\cite{zhu2017unpaired}}, with a total number of $6287$ images.
The test images are collected from \textit{Pexels}\footnote{\url{https://www.pexels.com/}} using keywords \textit{landscape} and \textit{nature}.
We prepare $974$ natural photos in total as input content images.

\paragraph{Baseline methods.}
To demonstrate the controllability and diversity of our method, we compare with recent 7 style transfer methods (i.e., Gatys~\cite{gatys2016neural}, AdaIN~\cite{huang2017arbitrary}, WCT~\cite{li2017universal}, AvatarNet~\cite{sheng2018avatar}, AAMS~\cite{yao2019attention}, SANet~\cite{park2019arbitrary}, and LinearST~\cite{li2018learning}), as well as 5 image-to-image translation methods (i.e., CycleGAN~\cite{zhu2017unpaired}, MUNIT~\cite{huang2018multimodal}, DRIT~\cite{lee2018diverse}, StarGAN~\cite{choi2018stargan}, and UFDN~\cite{liu2018unified}).
For a fair comparison, we use author released source code whenever possible and train all methods with default configurations on same training set with same number of iterations, except CycleGAN.
We train CycleGAN with dropout layers of probability 0.5 to make it to be feasible for multimodal image generation.
The new model is denoted as \textit{CycleGAN\_D}.
We evaluate the performance of all models on the same \textit{Pexels} test set as mentioned above.

\newcommand\multiguidancewidth{0.0963}
\newcommand\multiguidancewider{0.20}
\begin{figure*}
\centering
\begin{subfigure}{\multiguidancewidth\linewidth}
    \includegraphics[width=\linewidth,height=\linewidth]{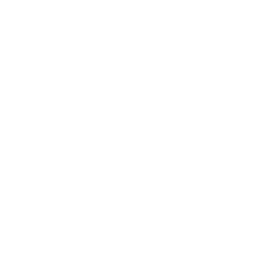}
    \includegraphics[width=\linewidth,height=\linewidth]{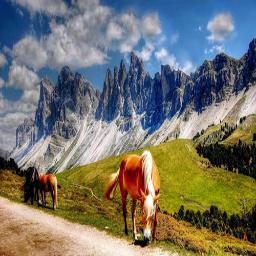}
    \includegraphics[width=\linewidth,height=\linewidth]{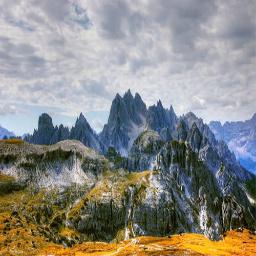}
    \includegraphics[width=\linewidth,height=\linewidth]{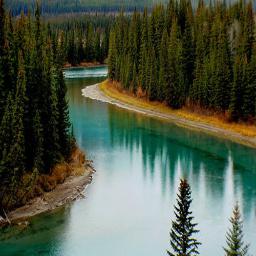}
    \includegraphics[width=\linewidth,height=\linewidth]{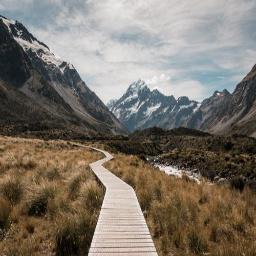}
\end{subfigure}
\begin{subfigure}{2pt}
    \rule{1pt}{265pt}%
\end{subfigure}
\begin{subfigure}{\multiguidancewidth\linewidth}
    \includegraphics[width=\linewidth,height=\linewidth]{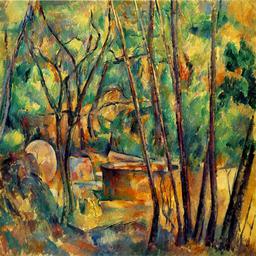}
    \includegraphics[width=\linewidth,height=\linewidth]{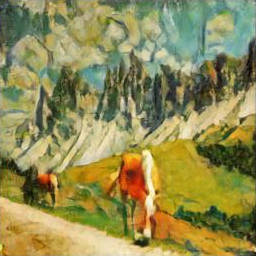}
    \includegraphics[width=\linewidth,height=\linewidth]{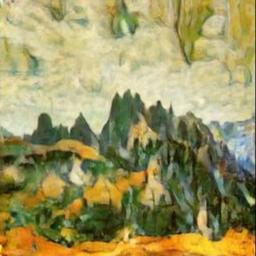}
    \includegraphics[width=\linewidth,height=\linewidth]{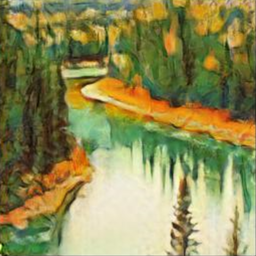}
    \includegraphics[width=\linewidth,height=\linewidth]{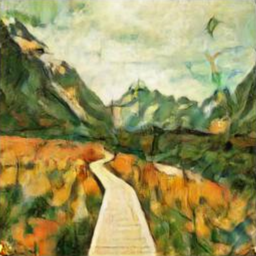}
\end{subfigure}
\begin{subfigure}{\multiguidancewidth\linewidth}
    \includegraphics[width=\linewidth,height=\linewidth]{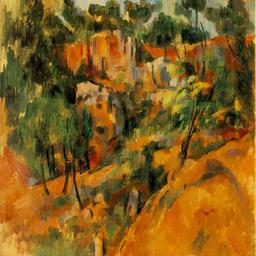}
    \includegraphics[width=\linewidth,height=\linewidth]{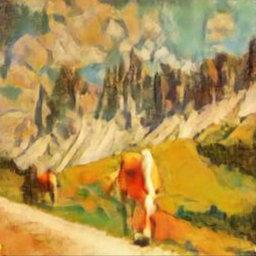}
    \includegraphics[width=\linewidth,height=\linewidth]{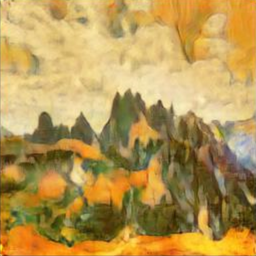}
    \includegraphics[width=\linewidth,height=\linewidth]{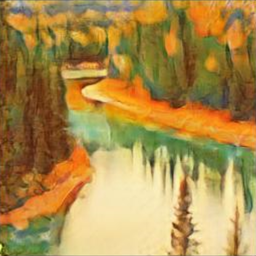}
    \includegraphics[width=\linewidth,height=\linewidth]{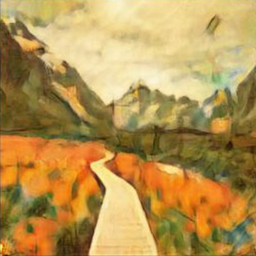}
\end{subfigure}
\begin{subfigure}{2pt}
    \rule{1pt}{265pt}%
\end{subfigure}
\begin{subfigure}{\multiguidancewidth\linewidth}
    \includegraphics[width=\linewidth,height=\linewidth]{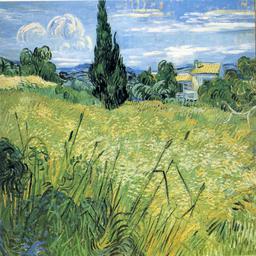}
    \includegraphics[width=\linewidth,height=\linewidth]{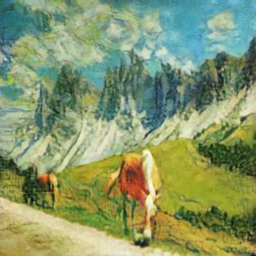}
    \includegraphics[width=\linewidth,height=\linewidth]{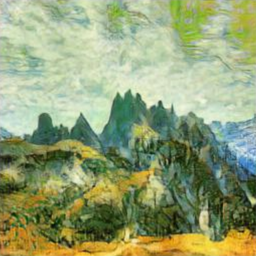}
    \includegraphics[width=\linewidth,height=\linewidth]{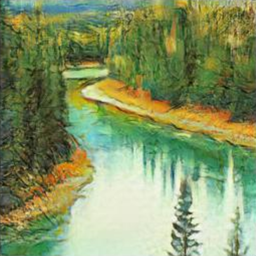}
    \includegraphics[width=\linewidth,height=\linewidth]{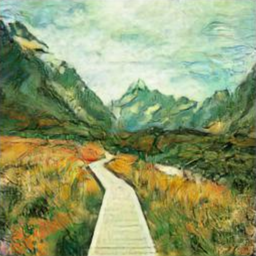}
\end{subfigure}
\begin{subfigure}{\multiguidancewidth\linewidth}
    \includegraphics[width=\linewidth,height=\linewidth]{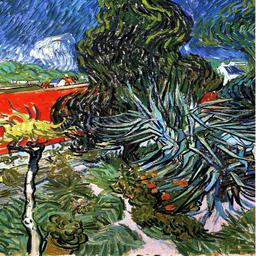}
    \includegraphics[width=\linewidth,height=\linewidth]{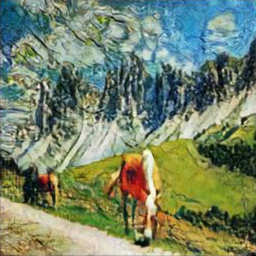}
    \includegraphics[width=\linewidth,height=\linewidth]{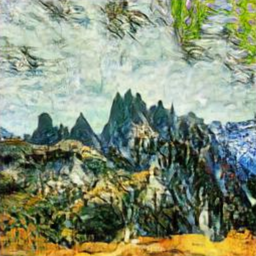}
    \includegraphics[width=\linewidth,height=\linewidth]{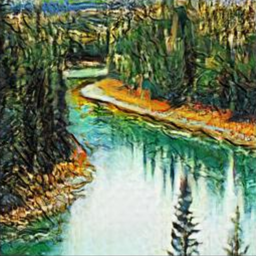}
    \includegraphics[width=\linewidth,height=\linewidth]{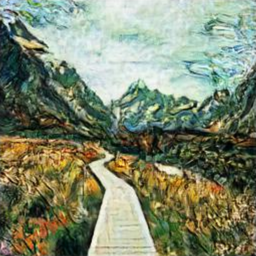}
\end{subfigure}
\begin{subfigure}{2pt}
    \rule{1pt}{265pt}%
\end{subfigure}
\begin{subfigure}{\multiguidancewidth\linewidth}
    \includegraphics[width=\linewidth,height=\linewidth]{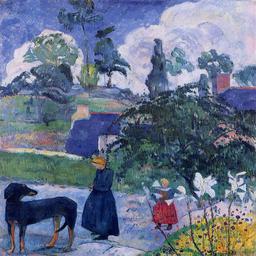}
    \includegraphics[width=\linewidth,height=\linewidth]{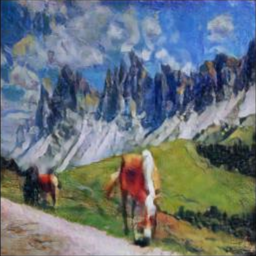}
    \includegraphics[width=\linewidth,height=\linewidth]{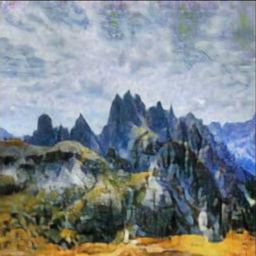}
    \includegraphics[width=\linewidth,height=\linewidth]{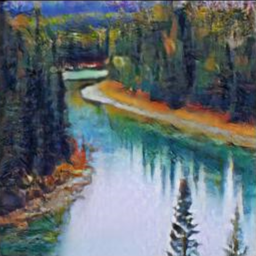}
    \includegraphics[width=\linewidth,height=\linewidth]{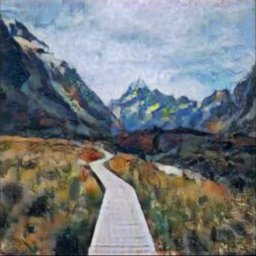}
\end{subfigure}
\begin{subfigure}{\multiguidancewidth\linewidth}
    \includegraphics[width=\linewidth,height=\linewidth]{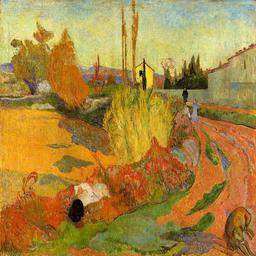}
    \includegraphics[width=\linewidth,height=\linewidth]{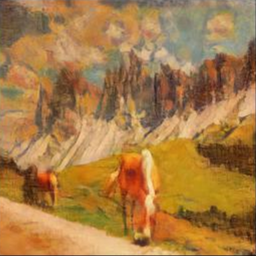}
    \includegraphics[width=\linewidth,height=\linewidth]{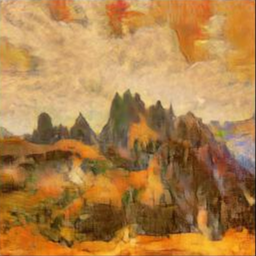}
    \includegraphics[width=\linewidth,height=\linewidth]{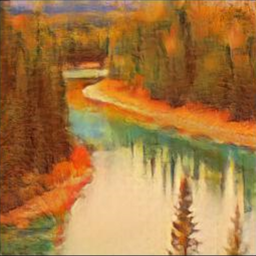}
    \includegraphics[width=\linewidth,height=\linewidth]{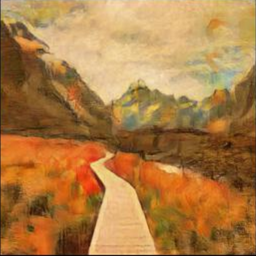}
\end{subfigure}
\begin{subfigure}{2pt}
    \rule{1pt}{265pt}%
\end{subfigure}
\begin{subfigure}{\multiguidancewidth\linewidth}
    \includegraphics[width=\linewidth,height=\linewidth]{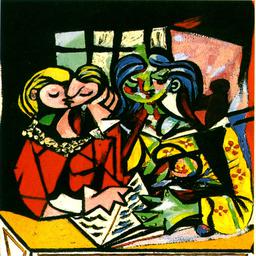}
    \includegraphics[width=\linewidth,height=\linewidth]{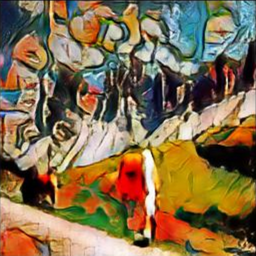}
    \includegraphics[width=\linewidth,height=\linewidth]{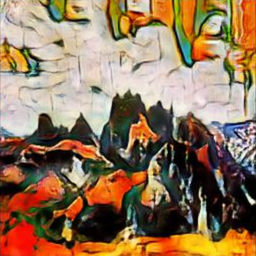}
    \includegraphics[width=\linewidth,height=\linewidth]{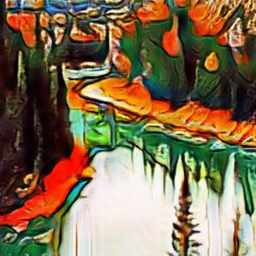}
    \includegraphics[width=\linewidth,height=\linewidth]{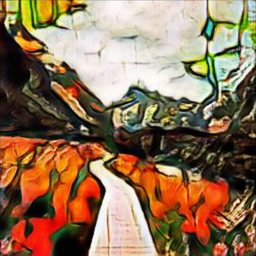}
\end{subfigure}
\begin{subfigure}{\multiguidancewidth\linewidth}
    \includegraphics[width=\linewidth,height=\linewidth]{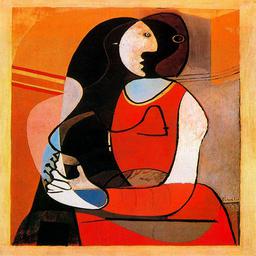}
    \includegraphics[width=\linewidth,height=\linewidth]{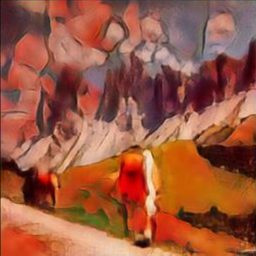}
    \includegraphics[width=\linewidth,height=\linewidth]{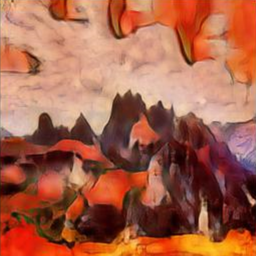}
    \includegraphics[width=\linewidth,height=\linewidth]{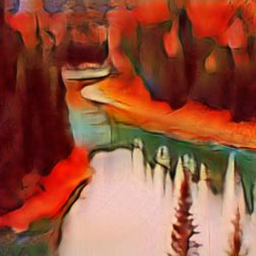}
    \includegraphics[width=\linewidth,height=\linewidth]{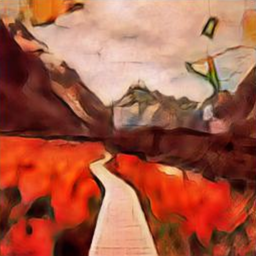}
\end{subfigure}

\begin{subfigure}{0.1\linewidth}
    \centering
    \caption{Photo}
\end{subfigure}
\begin{subfigure}{\multiguidancewider\linewidth}
    \centering
    \caption{C\'ezanne}
\end{subfigure}
\begin{subfigure}{\multiguidancewider\linewidth}
    \centering
    \caption{Van Gogh}
\end{subfigure}
\begin{subfigure}{\multiguidancewider\linewidth}
    \centering
    \caption{Gauguin}
\end{subfigure}
\begin{subfigure}{\multiguidancewider\linewidth}
    \centering
    \caption{Picasso}
\end{subfigure}

\caption{Style transfer results using artwork of different artists as exemplars. The first column shows input content images and the first row shows the guiding exemplars. For each artist (domain), two exemplars are used to demonstrate intra-domain discrepancy.}
\label{fig:guidance_multi_authors}
\vspace*{\afterimagevspace}
\end{figure*}

\subsection{Qualitative Evaluation}
\newcommand\changingartistswidth{0.32}
\begin{figure}
\centering
\begin{subfigure}{\changingartistswidth\linewidth}
    \includegraphics[width=\linewidth]{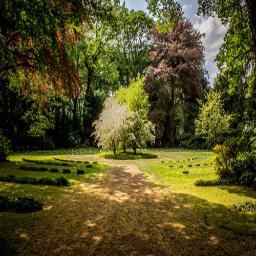}
    \caption{Photo}
\end{subfigure}
\begin{subfigure}{\changingartistswidth\linewidth}
    \includegraphics[width=\linewidth]{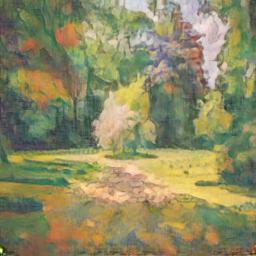}
    \caption{C\'ezanne}
\end{subfigure}
\begin{subfigure}{\changingartistswidth\linewidth}
    \includegraphics[width=\linewidth]{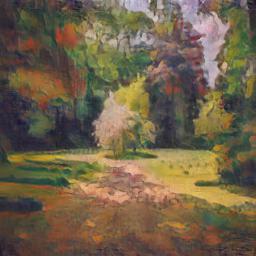}
    \caption{Gauguin}
\end{subfigure}
\vspace{3pt}
\begin{subfigure}{\changingartistswidth\linewidth}
    \includegraphics[width=\linewidth]{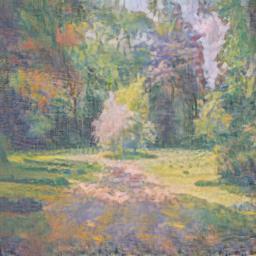}
    \caption{Monet}
\end{subfigure}
\begin{subfigure}{\changingartistswidth\linewidth}
    \includegraphics[width=\linewidth]{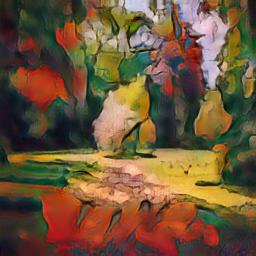}
    \caption{Picasso}
\end{subfigure}
\begin{subfigure}{\changingartistswidth\linewidth}
    \includegraphics[width=\linewidth]{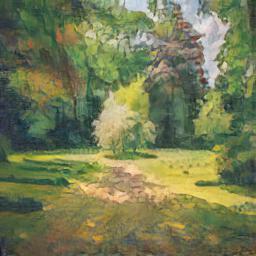}
    \caption{Van Gogh}
\end{subfigure}

\vspace{-7pt}
\caption{Results of changing artist domains. We fix the content image and style code to conduct style transfer with different artists' labels.}
\label{fig:artist_comparison}
\vspace*{\afterimagevspace}
\end{figure}

\newcommand\randominterpolationheight{0.11}
\newcommand\randominterpolationvspace{3pt}
\begin{figure*}
\centering
\includegraphics[height=\randominterpolationheight\linewidth]{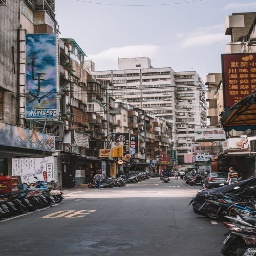}
\includegraphics[height=\randominterpolationheight\linewidth]{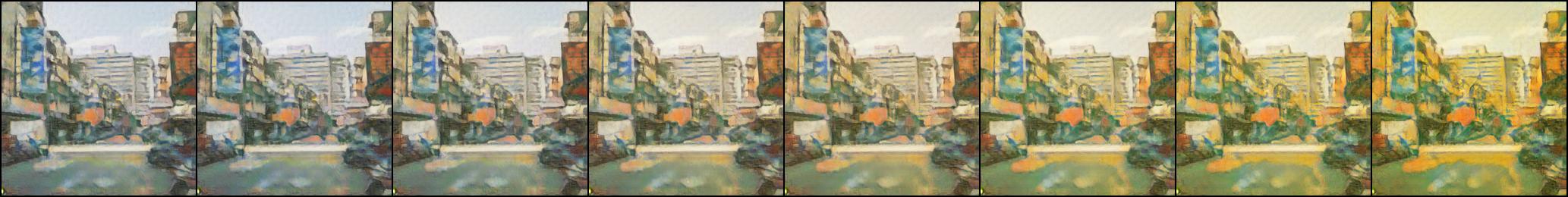}

\vspace{\randominterpolationvspace}
\includegraphics[height=\randominterpolationheight\linewidth]{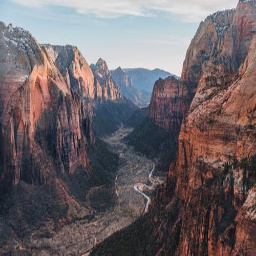}
\includegraphics[height=\randominterpolationheight\linewidth]{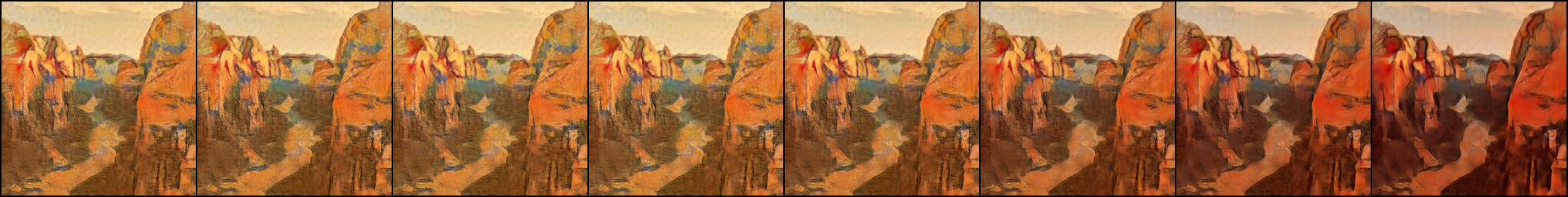}

\vspace{\randominterpolationvspace}
\includegraphics[height=\randominterpolationheight\linewidth]{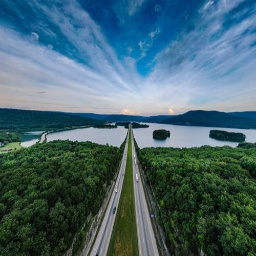}
\includegraphics[height=\randominterpolationheight\linewidth]{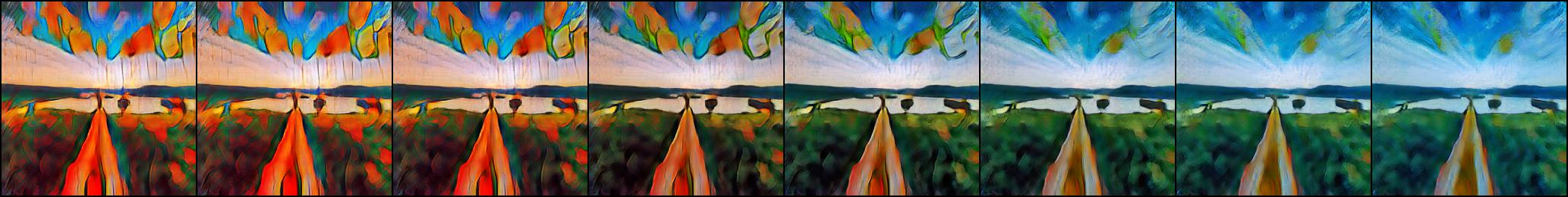}

\vspace{\randominterpolationvspace}
\includegraphics[height=\randominterpolationheight\linewidth]{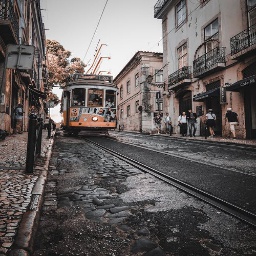}
\includegraphics[height=\randominterpolationheight\linewidth]{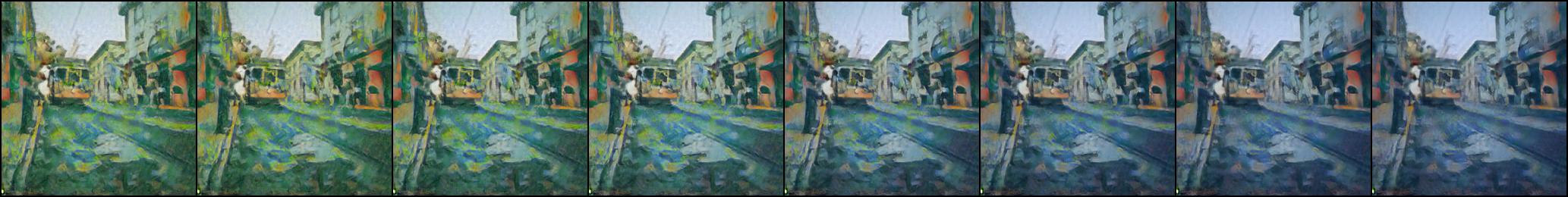}

\vspace{-7pt}
\caption{Linear interpolation results of two random styles. The content images are shown on the left, while two randomly sampled styles are shown in the second left-most and the right-most columns, respectively.}
\label{fig:random_interpolation}
\vspace*{\afterimagevspace}
\end{figure*}

\paragraph{Qualitative comparison.}
Figure~\ref{fig:guidance_comparison} shows the comparison of our exemplar-guided style transfer results with the ones of other approaches.
Each row corresponds to one artist's style (domain) and different columns represent different methods.
The corresponding content image and style image are shown in leftmost columns.
Overall, our method achieves more visually plausible results than others.
For example, Gatys et al.\commentout{~\cite{gatys2016neural}} (Figure~\ref{fig:guidance_comparison}c) fails to preserve content semantic information well and also to reproduce sky in first and third content image (1st and 3rd rows).
AdaIN\commentout{~\cite{huang2017arbitrary}} (Figure~\ref{fig:guidance_comparison}d) achieves high fidelity w.r.t input content images, but the results are often over-blurred.
WCT\commentout{~\cite{li2017universal}} (Figure~\ref{fig:guidance_comparison}e) cannot get satisfactory results with severely distorted contents and less consistent style w.r.t style exemplars.
AvatarNet\commentout{~\cite{sheng2018avatar}} (Figure~\ref{fig:guidance_comparison}f) and AAMS\commentout{~\cite{yao2019attention}} (Figure~\ref{fig:guidance_comparison}h) tend to generate either blurry results or images with granular artifacts.
MUNIT\commentout{~\cite{huang2018multimodal}} (Figure~\ref{fig:guidance_comparison}g) performs better than WCT, but also suffers blurring issues and dirty appearance artifacts (e.g. 2nd and 3rd rows).
Finally, while SANet (Figure~\ref{fig:guidance_comparison}i) and LinearST (Figure~\ref{fig:guidance_comparison}j) present balanced appearance between content and style, they still suffer from content distortions (Figure~\ref{fig:guidance_comparison}i, 2nd row) and color deviations (Figure~\ref{fig:guidance_comparison}j, 4th row).
Compared to these approaches, our results have less artifacts while achieve better visual quality, i.e., both content similarity and style consistency are well preserved.

\paragraph{Multimodal generation.}
Our method can generate diverse multimodal results in different ways for style guidance.
To demonstrate this advantage, for each domain representing the style of an artist, we generate multiple stylized images from the same content image (1) by random sampling in our learned style embedding space (see Figure~\ref{fig:sample_comparison}), and (2) by using difference reference images of the corresponding artist (see Figure~\ref{fig:guidance_multi_authors}).
In both cases, our method can generate vivid stylized images which are consistent with the unique style of each artist.
For example, the results guided by Picasso's artwork are composed of large color blocks in an abstract style, while the ones guided by Monet's work appear to be vague and are full of subtle strokes.

\paragraph{Multi-domain generation.}
Our style space decouples multi-domain control from multimodal generation. Figure~\ref{fig:artist_comparison} shows style transfer results with fixed style code and different artist domain labels.
In general, our method can generate images of different styles which preserve unique brush strokes and customary color collocations of each artist. For instance, the stylized photos from Picasso are more vivid and abstract while the ones from Van Gogh generates many tiny strokes.

\begin{figure}
\centering
\includegraphics[width=\linewidth]{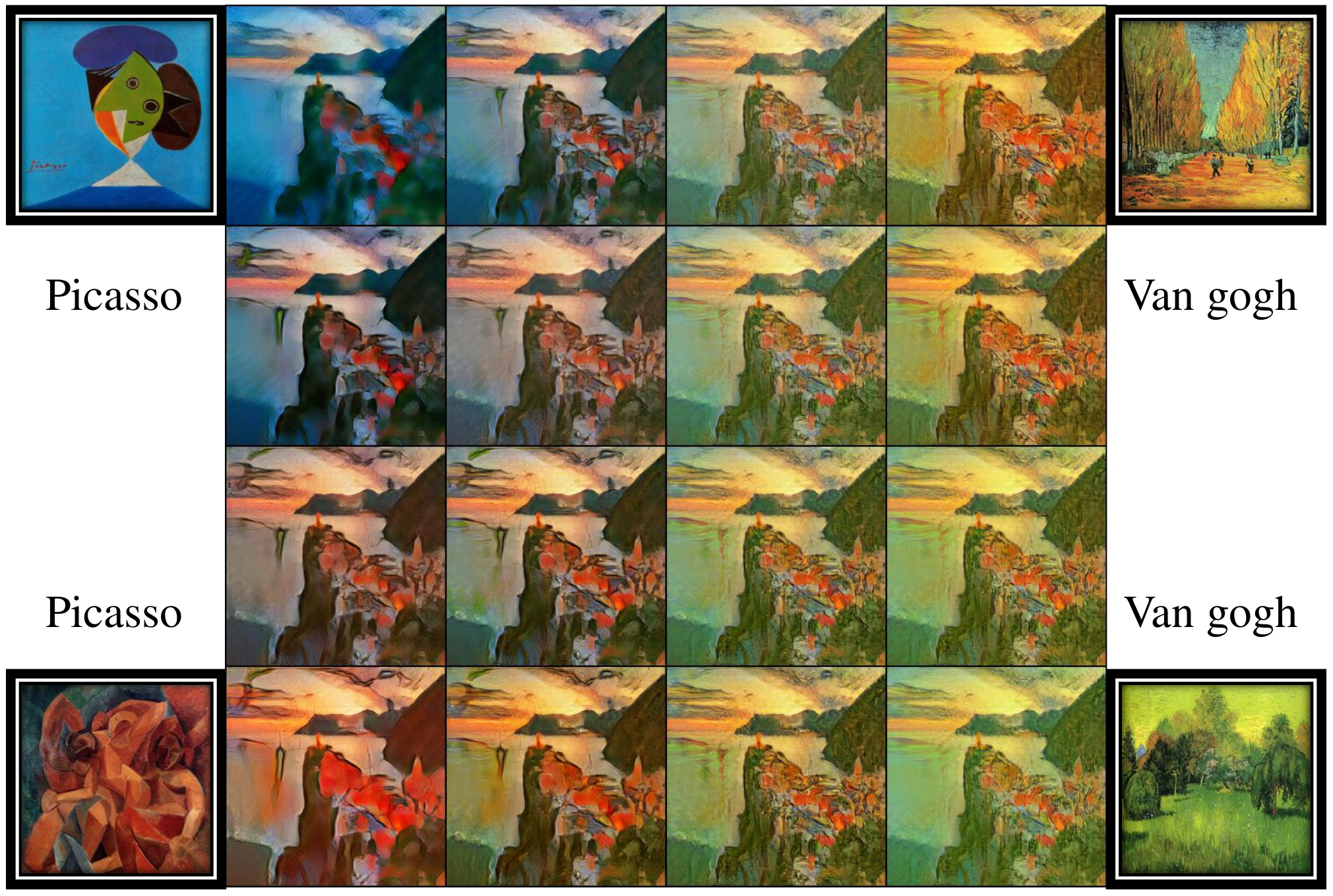}
\caption{Linear interpolation results of multiple styles. The input style images are shown in the four corners.}
\vspace*{\afterimagevspace}
\label{fig:interpolation}
\end{figure}

\paragraph{Style interpolation.}
To validate the smoothness of our latent style space, we present interpolation results using different guidance methods.
Figure~\ref{fig:random_interpolation} shows image sequences generated by linear interpolation of two randomly sampled style codes in latent space.
We can observe smooth and plausible style change as the interpolate weight varies.
Figure~\ref{fig:interpolation} demonstrates interpolation results between multiple styles defined using reference images shown in the four corners.
We obtain satisfactory intra-domain (vertically) and inter-domain (horizontally) interpolation results.

\subsection{Quantitative Evaluation}
\begin{table*}
\centering
\resizebox{\textwidth}{!}{
\begin{tabular}{|l*{11}{|c}|}
\hline
\multirow{2}{*}{Method} &
\multicolumn{2}{c|}{Photo2Monet (P2M)} &
\multicolumn{2}{c|}{Photo2Van Gogh (P2V)} &
\multicolumn{2}{c|}{Photo2C$\acute{e}$zanne (P2C)} &
\multicolumn{2}{c|}{Photo2Gauguin (P2G)} &
\multicolumn{2}{c|}{Photo2Picasso (P2P)} \\
\cline{2-11} & IS & LPIPS & IS & LPIPS & IS & LPIPS & IS & LPIPS & IS & LPIPS \\
\hline
\hline
CycleGAN\_D~\cite{zhu2017unpaired} &
$3.77\pm0.06$ & $0.16\pm0.05$ & $2.09\pm0.02$ & $0.16\pm0.05$ & $\mathbf{1.71\pm0.03}$ & $0.18\pm0.05$ & $2.04\pm0.03$ & $0.19\pm0.04$ & $1.70\pm0.02$ & $0.23\pm0.05$ \\
\hline
StarGAN~\cite{choi2018stargan} &
$2.64\pm0.26$ & $-$ & $1.72\pm0.03$ & $-$ & $1.51\pm0.00$ & $-$ & $1.46\pm0.01$ & $-$ & $1.22\pm0.01$ & $-$ \\
\hline
UFDN~\cite{liu2018unified} &
$2.35\pm0.02$ & $-$ & $1.93\pm0.03$ & $-$ & $1.36\pm0.01$ & $-$ & $1.49\pm0.01$ & $-$ & $1.33\pm0.01$ & $-$ \\
\hline
DRIT~\cite{lee2018diverse} &
$3.50\pm0.06$ & $0.30\pm0.12$ & $1.97\pm0.03$ & $0.32\pm0.12$ & $1.50\pm0.02$ & $0.23\pm0.09$ & $1.97\pm0.03$ & $0.24\pm0.10$ & $1.56\pm0.01$ & $0.25\pm0.10$ \\
\hline
MUNIT~\cite{huang2018multimodal} &
$\mathbf{4.25\pm0.10}$ & $0.44\pm0.12$ & $2.23\pm0.03$ & $0.46\pm0.13$ & $1.67\pm0.02$ & $0.37\pm0.11$ & $2.01\pm0.03$ & $0.39\pm0.11$ & $1.81\pm0.02$ & $\mathbf{0.54\pm0.12}$ \\
\hline
\hline
Ours+Exemplar &
$2.87\pm0.04$ & $0.21\pm0.11$ & $2.11\pm0.02$ & $0.30\pm0.13$ & $1.50\pm0.02$ & $0.19\pm0.09$ & $2.22\pm0.03$ & $0.20\pm0.10$ & $1.68\pm0.04$ & $0.31\pm0.12$ \\
\hline
Ours+Sample &
$3.62\pm0.04$ & $\mathbf{0.48\pm0.19}$ & $\mathbf{2.30\pm0.04}$ & $\mathbf{0.50\pm0.19}$ & $1.50\pm0.03$ & $\mathbf{0.43\pm0.17}$ & $\mathbf{2.39\pm0.04}$ & $\mathbf{0.45\pm0.17}$ & $\mathbf{1.88\pm0.03}$ &
$0.49\pm0.17$ \\
\hline
\hline
Real &
$7.01\pm1.58$ & $-$ & $2.53\pm0.61$ & $-$ & $3.77\pm1.23$ & $-$ & $2.58\pm0.44$ & $-$ & $2.92\pm0.85$ & $-$ \\
\hline
\end{tabular}}
\caption{Comparisons of IS and LPIPS scores of different stylization methods for each painter domain. One hundred content images are respectively stylized 100 times at random for calculating IS score. The LPIPS metric is measured using 1900 pairs stylized images for the domain of each artist.  For both metrics, the higher is the better. The last row presents the results of real data.
}
\label{tab:scores}
\end{table*}

\begin{table}
\centering
\begin{tabular}{|l*{6}{|r}|}
\hline
Method & P2M & P2V & P2C & P2G & P2P & Overall \\
\hline
StarGAN& 0.46 & $\mathbf{0.88}$ & 0.14 & 0.15 & 0.11 & 0.348 \\
\hline
UFDN & 0.92 & 0.56 & 0.71 & 0.16 & 0.10 & 0.490 \\
\hline
\hline
Ours + Sample & 0.76 & 0.26 & 0.41 & 0.33 & 0.31 & 0.414 \\
\hline
Ours + Exemplar & $\mathbf{0.97}$ & 0.47 & $\mathbf{0.83}$ & $\mathbf{0.69}$ & $\mathbf{0.50}$ & $\mathbf{0.692}$ \\
\hline
\end{tabular}
\caption{The artist classification accuracy of methods with a unified generator. For each domain, a hundred test images from \emph{Pexels.com} are used to perform stylization. The results are categorized by a ResNet-18 model finetuned on the test set.}
\vspace{-15pt}
\label{tab:accuracy}
\end{table}


To evaluate our stylized image quality, we conduct quantitative comparison using two different metrics, i.e., Learned Perceptual Image Patch Similarity (LPIPS) score and Inception score (IS).
The LPIPS metric~\cite{zhang2018unreasonable} is defined as a perceptual distance between image pairs and is calculated as a weighted difference between their embeddings on a pretrained VGG16 network, where the weights are fit so that the metric agrees with human perceptual similarity judgments.
The Inception score is the expectation of KL divergence distance between two sets of generated images, both in feature space extracted by a pretrained Inception-V3~\cite{szegedy2016rethinking} network, which measures generation quality and diversity.
To measure the quality for each artist style, we finetune the specific embedding network for each domain separately on our training data.
We select $100$ photos from test set and stylize them in specific domain with $100$ randomly sampled style codes for the IS metric and $19$ different style code pairs for the LPIPS metric.
We report the mean and standard variation of both LPIPS and IS in Table~\ref{tab:scores}.
As indicated by the scores, our method outperforms other methods in most test cases.
Specifically, \textit{CycleGAN\_D} gets the lowest LPIPS score and cannot generate diverse enough results, since the dropout layer only provides stochastic noise which leads to limited changes in the output.
StarGAN gets the lowest IS score, which indicates that it cannot generate high-quality stylized images.
We argue that the structure of StarGAN is designed for general multi-domain image translation and is not specifically optimized for the task of style transfer.
UFDN also does not perform well in most cases and fails to decopling content and style. Without any specific design and constrains, it is difficult to pass different style information into a single generator to synthesize plausible stylization results.
DRIT and MUNIT demonstrate better results than StarGAN and \textit{CycleGAN\_D} in terms of the two metrics.
Our method outperforms these methods in most cases.
The evaluation scores also show that our random-sample based generation is consistently better than exemplar-guided generation, which indicates that our random-sample approach is able to synthesize diverse and high-quality stylized images.

Finally, to demonstrate that our method preserves consistent domain information for multimodal generation, we conduct an experiment to re-classify stylized images into their corresponding style domains.
We train a classification network using style labels from our training data.
We compare our exemplar-guided and random-sample based approaches with two unified methods (i.e., StarGAN~\cite{huang2018multimodal} and UFDN~\cite{liu2018unified}) and report the corresponding results in Table~\ref{tab:accuracy}.
In most cases (4 out of 5), our method leads to higher classification accuracy, which indicates our output distribution is closer to the reference style.

\subsection{Ablation Study}

\newcommand\ablationstudywidth{0.24}
\begin{figure}
\centering
\begin{subfigure}{\ablationstudywidth\linewidth}
    \centering
    \includegraphics[width=\linewidth,height=\linewidth]{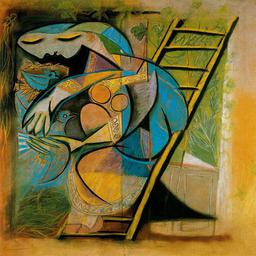}
    \caption{Style}
\end{subfigure}
\begin{subfigure}{\ablationstudywidth\linewidth}
    \includegraphics[height=\linewidth]{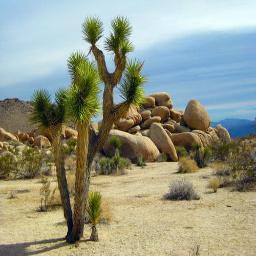}
    \caption{Photo}
\end{subfigure}
\begin{subfigure}{\ablationstudywidth\linewidth}
    \includegraphics[height=\linewidth]{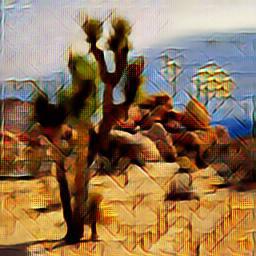}
    \caption{w/o ms $\contentdis$}
\end{subfigure}
\begin{subfigure}{\ablationstudywidth\linewidth}
    \includegraphics[height=\linewidth]{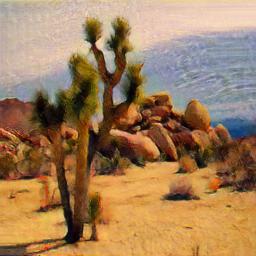}
    \caption{w/o $\conditionalidentityloss$}
\end{subfigure}

\hspace{\ablationstudywidth\linewidth}
\begin{subfigure}{\ablationstudywidth\linewidth}
    \includegraphics[height=\linewidth]{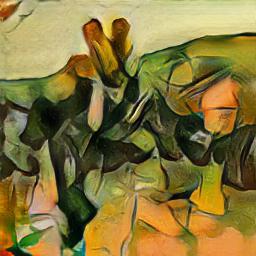}
    \caption{w/o $\contentpreservingloss$}
\end{subfigure}
\begin{subfigure}{\ablationstudywidth\linewidth}
    \includegraphics[height=\linewidth]{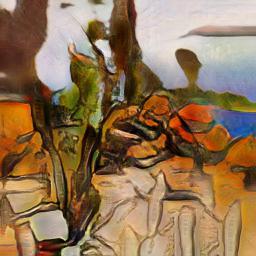}
    \caption{w/o $\stylepreservingloss$}
\end{subfigure}
\begin{subfigure}{\ablationstudywidth\linewidth}
    \includegraphics[height=\linewidth]{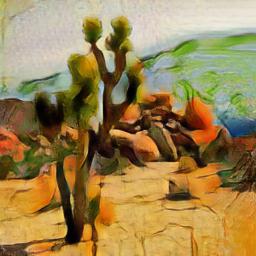}
    \caption{full model}
\end{subfigure}

\vspace{-7pt}
\caption{Qualitative results of ablation study. The influencing factors are respectively removed and the corresponding results are compared with our full model.}
\vspace*{\afterimagevspace}
\label{fig:ablation_visual}
\end{figure}

To analyze the effect of each component in our framework, we conduct a series of ablation study with certain components turned off and report the IS and LPIPS scores for evaluation.
Table~\ref{tab:ablation} shows the result of ablation study.
The first row shows the effect of replacing the multi-scale \mathenv{\contentdis} with a standard discriminator.
As shown, the corresponded model does not perform well either in terms of quality or diversity.
The result without $\stylepreservingloss$ shows a significant decrease on LPIPS score because there is no longer a requirement for style consistency.
Without $\contentpreservingloss$, the network no longer preserve the content of input reference, thus performs worst on both IS and LPIPS scores.
When $\conditionalidentityloss$ is turned off, we do observe a small increase in LPIPS score which indicates a slightly improved diversity.
However, the image quality decreases as indicated by decreased IS score.
Figure~\ref{fig:ablation_visual} demonstrates the visual quality under different ablation setups.
Our full model in Figure~\ref{fig:ablation_visual}g achieves the best trade-off between content fidelity and style consistency.

\subsection{User Study}
\label{sec:user_study}

\begin{table}
\centering
\begin{tabular}{|l*{2}{|c}|}
\hline
Method & IS  & LPIPS \\
\hline
\hline
Ours w/o ms \mathenv{\contentdis} & $2.17\pm0.03$ & $0.23\pm0.03$  \\
\hline
Ours w/o \mathenv{\stylepreservingloss} & $2.10\pm0.03$ & $0.13\pm0.02$  \\
\hline
Ours w/o \mathenv{\contentpreservingloss} &  $1.87\pm0.02$ & $0.11\pm0.02$ \\
\hline
Ours w/o \mathenv{\conditionalidentityloss} & $1.93\pm0.03$ & $0.65\pm0.02$ \\
\hline
Our full model & $2.34\pm0.04$ & $0.47\pm0.03$ \\
\hline
\end{tabular}
\caption{Ablation study of our method. Different factors are removed separately to evaluate the corresponding impact on the results.}
\vspace{-10pt}
\label{tab:ablation}
\end{table}


\commentout{
\begin{table}
\centering
 \begin{tabular}{|l|c|c|c|}
\hline
\multicolumn{1}{|c|}{\multirow{2}{*}{Methods}} & \multirow{2}{*}{\tabincell{c}{Mean of\\comparisons}} & \multicolumn{2}{c|}{\tabincell{c}{Mean of win rate}} \\ \cline{3-4}
\multicolumn{1}{|c|}{}               &                                      & Content fidelity       & Style preference       \\ \hline
\hline
 WCT~\cite{li2017universal}          &        $12.83$            &    $82.08\%$           &   $57.66\%$            \\ \hline
AdaIN~\cite{huang2017arbitrary}      &        $12.93$            &    $68.56\%$           &   $56.70\%$            \\ \hline
MUNIT~\cite{huang2018multimodal}     &        $13.13$            &    $81.73\%$           &   $68.53\%$            \\ \hline
AAMS~\cite{yao2019attention}         &        $13.10$            &    $83.21\%$           &   $75.57\%$            \\ \hline
SANet~\cite{park2019arbitrary}       &        $13.00$            &    $73.08\%$           &   $54.10\%$            \\ \hline \hline
Average                              &        $13.00$            &    $77.74\%$           &   $62.56\%$  \\ \hline
\end{tabular}
\caption{Results of our user study. The first column indicates the type of the contrast methods. The second column shows the average number of our
method compared with the others. We provide the average times that our results win the compared methods in terms of both content fidelity and style preference.}
\label{tab:user_study}
\end{table}
}

\begin{table}
\centering
\begin{tabular}{|l|c|c|c|}
\hline
\multicolumn{1}{|c|}{\multirow{2}{*}{Methods}} & \multicolumn{2}{c|}{\tabincell{c}{Mean of win rate}} \\ \cline{2-3}
\multicolumn{1}{|c|}{}               &  Content fidelity       & Style preference       \\ \hline
\hline
Ours vs WCT\toremove{~\cite{li2017universal} }         &            $82.08\%$           &   $57.66\%$            \\ \hline
Ours vs AdaIN\toremove{~\cite{huang2017arbitrary}}      &            $68.56\%$           &   $56.70\%$            \\ \hline
Ours vs MUNIT\toremove{~\cite{huang2018multimodal}}     &           $81.73\%$           &   $68.53\%$            \\ \hline
Ours vs AAMS\toremove{~\cite{yao2019attention}}         &           $83.21\%$           &   $75.57\%$            \\ \hline
Ours vs SANet\toremove{~\cite{park2019arbitrary}}       &            $73.08\%$           &   $54.10\%$            \\ \hline
\end{tabular}
\caption{Our user study results.
In each row, we report the average percentages that our results are selected when compare to these from the corresponding baseline method.
}
\label{tab:user_study}
\end{table}

To further evaluate our method quantitatively, we conduct user study to measure the preference of different stylization methods.
We select five exemplar-guided style transfer methods as baselines to compare, including WCT~\cite{li2017universal}, AdaIN~\cite{huang2017arbitrary}, MUNIT~\cite{huang2018multimodal}, AAMS~\cite{yao2019attention}, and SANet~\cite{park2019arbitrary}.
We hire $30$ annotators of different background and from different regions to answer $65$ randomly generated questions.
Given the content image and the exemplar, we show the subjects two stylization results in random order, one by our method and the other from a baseline approach.
For each pair of stylized images, the annotator is asked to answer two questions: (1) which one has higher fidelity to the content image; and (2) which one has the preferred style.
Our user study results are reported in Table~\ref{tab:user_study}.
In each row, we provide the average percentages that our results are selected when compare to those from the corresponding baseline method by the two aformentioned questions.
As shown in Table~\ref{tab:user_study}, our results are more preferred than these from baseline approaches by both measures.

\section{Conclusion}
In this paper, we present a new framework to multimodal and achieve multi-domain style transfer. 
To enable image stylization via both exemplar-based and randomly sampled guidance, we propose a novel style alignment module to construct an embedding style space.
The constructed space eliminates the explicit distribution gaps among various style domains and enables both image-guided style feature extraction and random style code generation, while reducing the risk of mode collapse due to the improper constraint by using KL-divergence loss.
Our framework shows superior performance for tasks of multimodal and multi-domain style transfer.
Extensive qualitative and quantitative evaluations demonstrate that our method outperforms previous style transfer methods.

\bibliographystyle{Transactions-Bibliography/IEEEtrans.bst}
\bibliography{main}

\begin{thebibliography}{10}
\providecommand{\url}[1]{#1}
\csname url@samestyle\endcsname
\providecommand{\newblock}{\relax}
\providecommand{\bibinfo}[2]{#2}
\providecommand{\BIBentrySTDinterwordspacing}{\spaceskip=0pt\relax}
\providecommand{\BIBentryALTinterwordstretchfactor}{4}
\providecommand{\BIBentryALTinterwordspacing}{\spaceskip=\fontdimen2\font plus
\BIBentryALTinterwordstretchfactor\fontdimen3\font minus
  \fontdimen4\font\relax}
\providecommand{\BIBforeignlanguage}[2]{{%
\expandafter\ifx\csname l@#1\endcsname\relax
\typeout{** WARNING: IEEEtranS.bst: No hyphenation pattern has been}%
\typeout{** loaded for the language `#1'. Using the pattern for}%
\typeout{** the default language instead.}%
\else
\language=\csname l@#1\endcsname
\fi
#2}}
\providecommand{\BIBdecl}{\relax}
\BIBdecl

\bibitem{anoosheh2018combogan}
A.~Anoosheh, E.~Agustsson, R.~Timofte, and L.~Van~Gool, ``{ComboGAN}:
  Unrestrained scalability for image domain translation,'' in \emph{Proceedings
  of the IEEE Conference on Computer Vision and Pattern Recognition Workshops},
  2018.

\bibitem{ba2016layer}
J.~L. Ba, J.~R. Kiros, and G.~E. Hinton, ``Layer normalization,'' \emph{arXiv
  preprint arXiv:1607.06450}, 2016.

\bibitem{bao2017cvae}
J.~Bao, D.~Chen, F.~Wen, H.~Li, and G.~Hua, ``{CVAE-GAN}: fine-grained image
  generation through asymmetric training,'' in \emph{IEEE International
  Conference on Computer Vision}, 2017.

\bibitem{chang2019sym}
S.~Chang, S.~Park, J.~Yang, and N.~Kwak, ``Sym-parameterized dynamic inference
  for mixed-domain image translation,'' in \emph{Proceedings of the IEEE
  International Conference on Computer Vision}, 2019, pp. 4803--4811.

\bibitem{choi2018stargan}
Y.~Choi, M.~Choi, M.~Kim, J.-W. Ha, S.~Kim, and J.~Choo, ``{StarGAN}: Unified
  generative adversarial networks for multi-domain image-to-image
  translation,'' in \emph{IEEE Conference on Computer Vision and Pattern
  Recognition}, 2018.

\bibitem{gatys2016neural}
L.~A. Gatys, A.~S. Ecker, and M.~Bethge, ``Image style transfer using
  convolutional neural networks,'' in \emph{IEEE Conference on Computer Vision
  and Pattern Recognition (CVPR)}, 2016, pp. 2414--2423.

\bibitem{huang2017arbitrary}
X.~Huang and S.~Belongie, ``Arbitrary style transfer in real-time with adaptive
  instance normalization,'' in \emph{Proceedings of the IEEE International
  Conference on Computer Vision}, 2017.

\bibitem{huang2018multimodal}
X.~Huang, M.-Y. Liu, S.~Belongie, and J.~Kautz, ``Multimodal unsupervised
  image-to-image translation,'' in \emph{Proceedings of the European Conference
  on Computer Vision (ECCV)}, 2018.

\bibitem{isola2017image}
P.~Isola, J.-Y. Zhu, T.~Zhou, and A.~A. Efros, ``Image-to-image translation
  with conditional adversarial networks,'' in \emph{IEEE Conference on Computer
  Vision and Pattern Recognition}, 2017.

\bibitem{kingma2014adam}
D.~P. Kingma and J.~Ba, ``Adam: A method for stochastic optimization,''
  \emph{arXiv preprint arXiv:1412.6980}, 2014.

\bibitem{kingma2013auto}
D.~P. Kingma and M.~Welling, ``Auto-encoding variational bayes,'' \emph{arXiv
  preprint arXiv:1312.6114}, 2013.

\bibitem{kotovenko2019contentandstyle}
D.~Kotovenko, A.~Sanakoyeu, S.~Lang, and B.~Ommer, ``Content and style
  disentanglement for artistic style transfer,'' in \emph{Proceedings of the
  IEEE International Conference on Computer Vision}, 2019.

\bibitem{kotovenko2019content}
D.~Kotovenko, A.~Sanakoyeu, P.~Ma, S.~Lang, and B.~Ommer, ``A content
  transformation block for image style transfer,'' in \emph{Proceedings of the
  IEEE Conference on Computer Vision and Pattern Recognition}, 2019, pp.
  10\,032--10\,041.

\bibitem{lee2018diverse}
H.-Y. Lee, H.-Y. Tseng, J.-B. Huang, M.~Singh, and M.-H. Yang, ``Diverse
  image-to-image translation via disentangled representations,'' in
  \emph{Proceedings of the European Conference on Computer Vision (ECCV)},
  2018.

\bibitem{lee2020drit++}
H.-Y. Lee, H.-Y. Tseng, Q.~Mao, J.-B. Huang, Y.-D. Lu, M.~Singh, and M.-H.
  Yang, ``Drit++: Diverse image-to-image translation via disentangled
  representations,'' \emph{International Journal of Computer Vision}, pp.
  1--16, 2020.

\bibitem{li2018learning}
X.~Li, S.~Liu, J.~Kautz, and M.-H. Yang, ``Learning linear transformations for
  fast arbitrary style transfer,'' \emph{arXiv preprint arXiv:1808.04537},
  2018.

\bibitem{li2017universal}
Y.~Li, C.~Fang, J.~Yang, Z.~Wang, X.~Lu, and M.-H. Yang, ``Universal style
  transfer via feature transforms,'' in \emph{Advances in Neural Information
  Processing Systems}, 2017, pp. 386--396.

\bibitem{liu2018unified}
A.~H. Liu, Y.-C. Liu, Y.-Y. Yeh, and Y.-C.~F. Wang, ``A unified feature
  disentangler for multi-domain image translation and manipulation,'' in
  \emph{Advances in Neural Information Processing Systems}, 2018.

\bibitem{liu2017unsupervised}
M.-Y. Liu, T.~Breuel, and J.~Kautz, ``Unsupervised image-to-image translation
  networks,'' in \emph{Advances in Neural Information Processing Systems},
  2017.

\bibitem{liu2019few}
M.-Y. Liu, X.~Huang, A.~Mallya, T.~Karras, T.~Aila, J.~Lehtinen, and J.~Kautz,
  ``Few-shot unsupervised image-to-image translation,'' in \emph{Proceedings of
  the IEEE International Conference on Computer Vision}, 2019, pp.
  10\,551--10\,560.

\bibitem{liu2020gmm}
Y.~Liu, M.~De~Nadai, J.~Yao, N.~Sebe, B.~Lepri, and X.~Alameda-Pineda,
  ``Gmm-unit: Unsupervised multi-domain and multi-modal image-to-image
  translation via attribute gaussian mixture modeling,'' \emph{arXiv preprint
  arXiv:2003.06788}, 2020.

\bibitem{ma2019exemplar}
L.~Ma, X.~Jia, S.~Georgoulis, T.~Tuytelaars, and L.~Van~Gool, ``Exemplar guided
  unsupervised image-to-image translation with semantic consistency,''
  \emph{Proceedings of ICLR}, 2019.

\bibitem{makhzani2015adversarial}
A.~Makhzani, J.~Shlens, N.~Jaitly, I.~Goodfellow, and B.~Frey, ``Adversarial
  autoencoders,'' \emph{arXiv preprint arXiv:1511.05644}, 2015.

\bibitem{mao2019mode}
Q.~Mao, H.-Y. Lee, H.-Y. Tseng, S.~Ma, and M.-H. Yang, ``Mode seeking
  generative adversarial networks for diverse image synthesis,'' in \emph{IEEE
  Conference on Computer Vision and Pattern Recognition}, 2019.

\bibitem{mao2017least}
X.~Mao, Q.~Li, H.~Xie, R.~Y. Lau, Z.~Wang, and S.~Paul~Smolley, ``Least squares
  generative adversarial networks,'' in \emph{Proceedings of the IEEE
  International Conference on Computer Vision}, 2017.

\bibitem{mirza2014conditional}
M.~Mirza and S.~Osindero, ``Conditional generative adversarial nets,''
  \emph{arXiv preprint arXiv:1411.1784}, 2014.

\bibitem{odena2017conditional}
A.~Odena, C.~Olah, and J.~Shlens, ``Conditional image synthesis with auxiliary
  classifier gans,'' in \emph{Proceedings of the 34th International Conference
  on Machine Learning-Volume 70}.\hskip 1em plus 0.5em minus 0.4em\relax JMLR.
  org, 2017, pp. 2642--2651.

\bibitem{park2019arbitrary}
D.~Y. Park and K.~H. Lee, ``Arbitrary style transfer with style-attentional
  networks,'' in \emph{Proceedings of the IEEE Conference on Computer Vision
  and Pattern Recognition}, 2019, pp. 5880--5888.

\bibitem{romero2019smit}
A.~Romero, P.~Arbel{\'a}ez, L.~Van~Gool, and R.~Timofte, ``{SMIT}: Stochastic
  multi-label image-to-image translation,'' in \emph{IEEE International
  Conference on Computer Vision Workshops}, 2019.

\bibitem{sanakoyeu2018style}
A.~Sanakoyeu, D.~Kotovenko, S.~Lang, and B.~Ommer, ``A style-aware content loss
  for real-time hd style transfer,'' in \emph{Proceedings of the European
  Conference on Computer Vision (ECCV)}, 2018.

\bibitem{sheng2018avatar}
L.~Sheng, Z.~Lin, J.~Shao, and X.~Wang, ``Avatar-net: Multi-scale zero-shot
  style transfer by feature decoration,'' in \emph{Proceedings of the IEEE
  Conference on Computer Vision and Pattern Recognition}, 2018, pp. 8242--8250.

\bibitem{sohn2015learning}
K.~Sohn, H.~Lee, and X.~Yan, ``Learning structured output representation using
  deep conditional generative models,'' in \emph{Advances in Neural Information
  Processing Systems}, 2015.

\bibitem{szegedy2016rethinking}
C.~Szegedy, V.~Vanhoucke, S.~Ioffe, J.~Shlens, and Z.~Wojna, ``Rethinking the
  inception architecture for computer vision,'' in \emph{Proceedings of the
  IEEE Conference on Computer Vision and Pattern Recognition}, 2016.

\bibitem{ulyanov2016instance}
D.~Ulyanov, A.~Vedaldi, and V.~Lempitsky, ``Instance normalization: The missing
  ingredient for fast stylization,'' \emph{arXiv preprint arXiv:1607.08022},
  2016.

\bibitem{viazovetskyi2020stylegan2}
Y.~Viazovetskyi, V.~Ivashkin, and E.~Kashin, ``Stylegan2 distillation for
  feed-forward image manipulation,'' \emph{arXiv preprint arXiv:2003.03581},
  2020.

\bibitem{yang2019multi}
F.-E. Yang, J.-C. Chang, C.-C. Tsai, and Y.-C.~F. Wang, ``A multi-domain and
  multi-modal representation disentangler for cross-domain image manipulation
  and classification,'' \emph{IEEE Transactions on Image Processing}, 2019.

\bibitem{yao2019attention}
Y.~Yao, J.~Ren, X.~Xie, W.~Liu, Y.-J. Liu, and J.~Wang, ``Attention-aware
  multi-stroke style transfer,'' in \emph{Proceedings of the IEEE Conference on
  Computer Vision and Pattern Recognition}, 2019, pp. 1467--1475.

\bibitem{yu2019multi}
X.~Yu, Y.~Chen, S.~Liu, T.~Li, and G.~Li, ``Multi-mapping image-to-image
  translation via learning disentanglement,'' in \emph{Advances in Neural
  Information Processing Systems}, 2019.

\bibitem{zhang2018unreasonable}
R.~Zhang, P.~Isola, A.~A. Efros, E.~Shechtman, and O.~Wang, ``The unreasonable
  effectiveness of deep features as a perceptual metric,'' in \emph{Proceedings
  of the IEEE Conference on Computer Vision and Pattern Recognition}, 2018.

\bibitem{zhu2017unpaired}
J.-Y. Zhu, T.~Park, P.~Isola, and A.~A. Efros, ``Unpaired image-to-image
  translation using cycle-consistent adversarial networks,'' in \emph{IEEE
  International Conference on Computer Vision}, 2017.

\end{thebibliography}

\end{document}